\documentclass{article}

\usepackage[preprint]{neurips_2026}

\usepackage[utf8]{inputenc} 
\usepackage[T1]{fontenc}    
\usepackage{hyperref}       
\usepackage{url}            
\usepackage{booktabs}       
\usepackage{amsfonts}       
\usepackage{amssymb}        
\usepackage{nicefrac}       
\usepackage{microtype}      
\usepackage[table]{xcolor}         
\usepackage{colortbl} 
\usepackage{multirow} 
\usepackage{makecell} 
\usepackage{graphicx} 
\usepackage{adjustbox} 
\usepackage{wrapfig} 
\usepackage{amsmath}

\definecolor{ourscolor}{RGB}{230,245,255}

\title{DSAQuant: Denoising-Stage-Aligned Quantization-Aware Training for Video Generation}

\author{
\textbf{Shuaiting Li}\textsuperscript{1,2}
\qquad
\textbf{Zelin Gao}\textsuperscript{1,4}
\qquad
\textbf{Haibin Shen}\textsuperscript{2}
\qquad
\textbf{Yujun Shen}\textsuperscript{1}
\\[0.4em]
\textbf{Haotong Qin}\textsuperscript{3$\dagger$}
\qquad
\textbf{Yinghao Xu}\textsuperscript{4,1$\dagger$}
\\[0.7em]
\textsuperscript{1}Robbyant
\qquad
\textsuperscript{2}ZJU
\qquad
\textsuperscript{3}PolyU
\qquad
\textsuperscript{4}HKUST
\\[0.5em]
\textsuperscript{$\dagger$}Corresponding Authors
}

\begin{document}

\maketitle

\begin{abstract}
\vspace{-0.5em}
Video diffusion models (VDMs) have achieved impressive progress in text-to-video generation, but their high memory and computational costs hinder practical deployment. Quantization-aware training (QAT) is an effective solution for compressing and accelerating advanced generative models without runtime overhead at inference.
However, existing QAT methods suffer from a distinctive challenge in VDMs: while they often preserve prompt semantics, global layout, and coarse motion, the quantized model severely degrades visual details, texture fidelity, and sharpness. In this paper, we trace this degradation to the timestep-agnostic design of conventional quantization pipelines, which overlooks the stage-wise functionality of video denoising. In VDMs, early denoising steps mainly establish global structure and motion, whereas middle and late steps refine local appearance and high-frequency details.
Based on this insight, we propose \textbf{DSAQuant}, a \textbf{D}enoising-\textbf{S}tage-\textbf{A}ligned \textbf{Quant}ization-aware training framework for VDMs.
During training, \textit{Denoising-Stage Oriented Supervision} preserves teacher distillation in early steps for stable structure planning, while shifting later steps toward target-driven optimization to enhance detail reconstruction.
During inference, \textit{Denoising-Stage Gated Guidance} disables CFG in the final denoising steps to prevent it from amplifying quantization-induced errors into high-frequency artifacts.
Extensive experiments on the Wan and CogVideoX families under W4A4 and W3A3 settings show that DSAQuant consistently outperforms the SOTA QAT baseline, improving the VBench average score by up to 6.60 under aggressive W3A3 quantization while preserving strong text-video alignment.
These results demonstrate that effective VDM quantization requires not only reducing quantization error, but also aligning quantization training and inference with the stage-wise nature of video diffusion.
\end{abstract}

\begin{center}
    \textbf{Project page:} \url{https://robbyant-research.github.io/DSAQuant/} \\
    \textbf{Code:} \url{https://github.com/robbyant-research/DSAQuant}
\end{center}

\section{Introduction}

Video diffusion models (VDMs)~\citep{yang2025cogvideox, wanteam2025wan, kong2025hunyuanvideo} produce state-of-the-art text-to-video at compute and memory costs that block real deployment, making quantization a critical path to practical use. Existing post-training quantization (PTQ) methods~\citep{zhao2025viditq, li2025svdquant, ashkboos2024quarot} rely on online transformations whose per-layer runtime overhead on high-dimensional video tokens largely offsets the low-bit speedup, leaving advanced VDMs with limited practical benefit from low-bit PTQ.

Quantization-aware training (QAT) sidesteps this overhead by directly optimizing the model under low-bit constraints, with no auxiliary activation transformation at inference time. However, the only existing QAT framework tailored for VDMs~\citep{huang2026qvgen} employs a timestep-agnostic MSE distillation pipeline, which applies the same teacher-imitation objective and the same classifier-free guidance (CFG) schedule across every denoising step. Yet video denoising itself is not timestep-agnostic: early steps establish global layout and motion, whereas middle and late steps refine textures and frame-level sharpness. Figure~\ref{fig:motivation} makes the consequence concrete: a conventional quantized VDM still follows the prompt and reproduces the global layout (Fig.~\ref{fig:motivation}a), yet its imaging-quality and sharpness scores collapse while prompt-alignment scores stay close to FP (Fig.~\ref{fig:motivation}a), and a stage-wise substitution experiment localizes the challenge to the middle and late timesteps (Fig.~\ref{fig:motivation}b). A timestep-agnostic quantization strategy is thus structurally mismatched to VDMs.

We trace this mismatch to two timestep-agnostic design choices. 
\textbf{On the training side}, a low-bit student forced to imitate the FP teacher in late steps may spend scarce discrete capacity matching the teacher's continuous-space fitting residual, without improving its own target prediction for detail reconstruction. 
\textbf{On the inference side}, the FP-level conditional and unconditional predictions converge in late steps, so the true guidance signal vanishes, while the two branches are quantized through independent activation grids whose errors are only weakly aligned and tend to accumulate rather than cancel; CFG then multiplicatively amplifies this uncorrelated quantization noise into the high-frequency artifacts we observe.

Building on this two-sided diagnosis, we propose \textbf{DSAQuant}, a denoising-stage-aligned QAT framework with two complementary components, one for each diagnosed failure. \textbf{Denoising-Stage Oriented Supervision} addresses the training-side capacity drain: we replace uniform MSE distillation with an exponential schedule that smoothly transfers supervision from a pure KD objective at $t=0$ to a pure target-loss objective at $t=1$. Early steps, therefore, inherit KD's variance reduction against the noisy STE gradient and stably anchor global structure, while late steps decouple the quantized student from the FP teacher's continuous-space fitting residual, freeing discrete capacity for genuine detail reconstruction. \textbf{Denoising-Stage Gated Guidance} addresses the inference-side noise amplification: we disable CFG in the final few denoising steps, removing the multiplicative amplifier that converts weakly aligned cond/uncond quantization errors into high-frequency artifacts, while preserving the strong text-video alignment that CFG provides at earlier steps. The two changes are orthogonal and complementary: the training-side change gives the quantized model the capacity to produce correct details, while the inference-side change ensures those details survive guidance. We validate DSAQuant across diverse VDMs (CogVideoX-2B/1.5-5B, Wan2.1-1.3B/14B, Wan2.2-5B) under both W4A4 and W3A3, achieving consistent and substantial improvements over the SOTA QAT baseline QVGen.

Our contributions are summarized as follows:
\begin{itemize}
\item We identify and empirically localize a \emph{stage-dependent} failure mode in quantized VDMs: timestep-agnostic quantization preserves prompt semantics but severely degrades the detail-refinement capability of middle and late denoising steps.
\item We propose \textbf{DSAQuant}, a denoising-stage-aligned QAT framework with two complementary components: \emph{Denoising-Stage Oriented Supervision} with an exponential schedule that decouples the quantized student from the FP teacher's continuous-space residual and frees discrete capacity for late-step detail; and \emph{Denoising-Stage Gated Guidance}, supported by a noise-amplification analysis showing that CFG multiplicatively amplifies the weakly aligned cond/uncond quantization errors in late timesteps.
\item Extensive experiments on diverse VDMs (CogVideoX-2B/1.5-5B, Wan2.1-1.3B/14B, Wan2.2-5B) under W4A4 and W3A3 settings demonstrate consistent and substantial improvements over SOTA QAT baselines, validating the effectiveness of our framework.
\end{itemize}

\section{Related Work}
\label{sec:related_work}

\textbf{Video diffusion models.}
Video Diffusion models (VDMs) based on diffusion transformer~\citep{peebles2023scalable} architectures have achieved remarkable success in high-quality video generation~\citep{yang2025cogvideox, wanteam2025wan, kong2025hunyuanvideo, opensora2024}.  However, advanced video DiTs often contain billions of parameters and require multi-step denoising over long spatio-temporal sequences, leading to substantial latency and memory costs. To improve efficiency, prior works have explored step distillation~\citep{yin2025causvid, lin2025diffusion}, sparse attention~\citep{xi2025sparsevideogen}, step caching~\citep{lv2025fastercache, zou2025accelerating, zhou2025less} and quantization~\citep{zhao2025viditq, feng2025s, huang2026qvgen}. 

\textbf{Quantization for diffusion models.}
Quantization~\citep{jacob2017quantization} reduces model size and inference cost by representing weights and activations with low-bit values. Most video diffusion quantization methods focus on post-training quantization (PTQ). Q-DiT~\citep{chen2024qdit} applies PTQ to VDMs with dynamic activation quantization, and ViDiT-Q~\citep{zhao2025viditq} further improves PTQ with mixed precision. S2Q-VDiT~\citep{feng2025s} leverages salient data and sparse token distillation. However, recent PTQ methods can still suffer from quality loss under activation quantization, and rotation-based transformations may introduce non-negligible runtime overhead in DiTs. QVGen~\citep{huang2026qvgen} first explored quantization-aware training (QAT) for VDMs, showing that previous QAT methods on image diffusion~\citep{he2024efficientdm, li2023qdm} can be applied to VDMs, and further improves weight quantization with auxiliary modules. We focus on the most relevant works in the main text and provide a more comprehensive discussion in Appendix~\ref{app:related_work}.

\section{Preliminary}
\subsection{Video Diffusion Sampling}
\label{subsec:vdm_sampling}
Video diffusion models (VDMs) generate videos through an iterative denoising process.  Let $x_1$ be the clean video latent and $\epsilon \sim \mathcal{N}(0,I)$ be Gaussian noise. We use $t \in [0,1]$ to denote the sampling progress from noise to data, and write the noising path as
\begin{equation}
    x_t = \alpha_t x_1 + \sigma_t \epsilon,
    \label{eq:path}
\end{equation}
where $\alpha_0=0$, $\sigma_0=1$, $\alpha_1=1$, and $\sigma_1=0$.
This notation includes DDIM-style sampling by choosing $\alpha_t=\sqrt{\bar{\alpha}_{\tau(t)}}$ and $\sigma_t=\sqrt{1-\bar{\alpha}_{\tau(t)}}$, and flow matching by choosing, e.g., $\alpha_t=t$ and $\sigma_t=1-t$.

Under this formulation, the original training target can be written generically as
\begin{equation}
    y_t = a_t x_1 + b_t \epsilon,
    \label{eq:true_target}
\end{equation}
At each step, the denoising network tries to predict a diffusion target $f(x_t,c,t)$, 
where the target can be noise, velocity, or a data-related quantity.

\begin{figure}[t]
    \centering
    \includegraphics[width=0.99\linewidth]{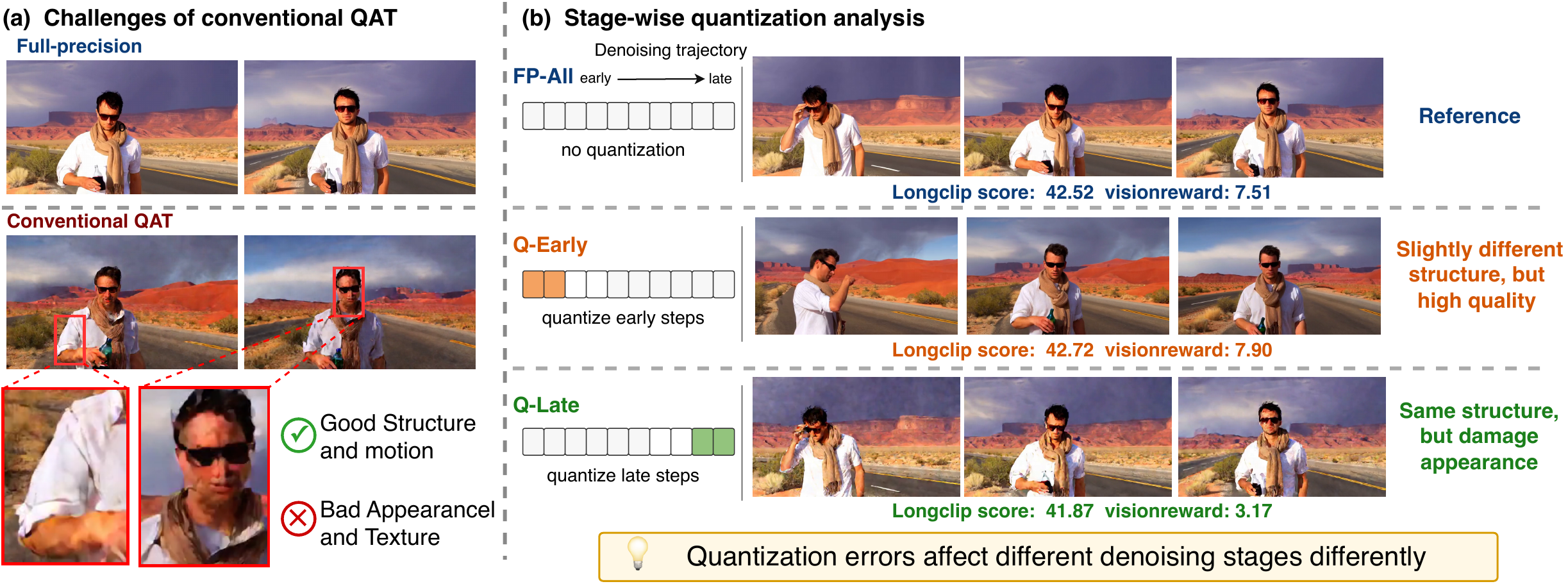}
    \caption{\textbf{Analysis of conventional QAT on VDMs (please zoom in for details)}. (a) Conventional methods generally preserve layout and motion, but suffer severe degradation in details and texture. (b) We apply quantization to different denoising stages of VDMs: quantizing early steps slightly perturbs layout but preserves perceptual quality, while quantizing late steps significantly damages visual quality.}
    \label{fig:motivation}
    \vspace{-1.0em}
\end{figure}

\subsection{Stage-Wise Quantization Analysis for Video Denoising}
\label{subsec:stage_behavior}

The denoising process of VDMs is stage-dependent: with signal-to-noise ratio $\mathrm{SNR}_t = \alpha_t^2 / \sigma_t^2$, early steps have low SNR and rely on the text condition and generative prior to establish global semantics, scene layout, and coarse motion, while late steps operate on latents that already carry recognizable structure and focus on refining local appearance, high-frequency textures, and frame-level sharpness.

Despite this stage-wise functionality, conventional quantization-aware training (QAT) treats every denoising step uniformly~\cite{he2024efficientdm, li2023qdm, huang2026qvgen}: it optimizes a quantized model $f^{q}$ to match the full-precision model $f^{\mathrm{fp}}$ across all steps indiscriminately. However, as shown in Figure~\ref{fig:motivation}(a), directly applying such uniform QAT to VDMs leads to a distinctive degradation pattern: the quantized model preserves prompt semantics, global layout, and coarse motion, but suffers from blurred textures, corrupted details, reduced sharpness, and unstable local motion.

To identify the source of this degradation, we perform a stage-wise quantization analysis (Figure~\ref{fig:motivation}(b)), in which different sampling stages are quantized separately while the remaining stages are kept full-precision. Quantizing only the early steps slightly perturbs global layout but preserves perceptual quality, while quantizing the middle and late steps reproduces the full failure mode. Consistent with the stage-wise picture above, conventional QAT fails not because it destroys semantic generation, but because its timestep-agnostic objective is inadequate for the detail-sensitive late stages. These observations motivate a denoising-stage-aligned framework that adapts both training supervision and inference guidance to the role each sampling stage plays.

\section{Method}
\begin{figure}
    \centering
    \includegraphics[width=0.99\linewidth]{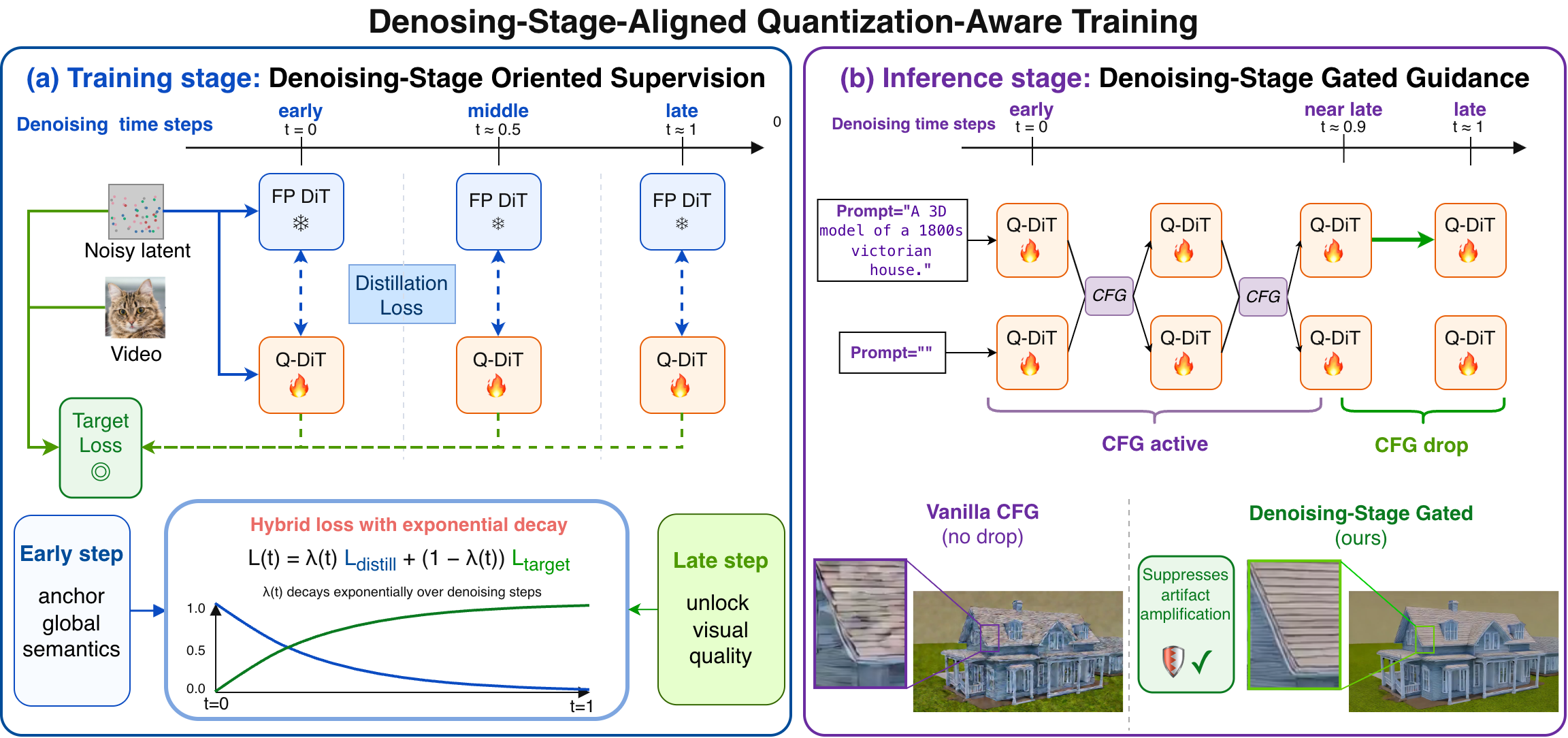}
    \caption{Overview of \textbf{DSAQuant}. (a) During training, \textbf{Denoising-Stage Oriented Supervision} smoothly transfers supervision from KD in early steps to the target loss in late steps via an exponential schedule, freeing late-step capacity for detail reconstruction. (b) During inference, \textbf{Denoising-Stage Gated Guidance} disables CFG in the final few steps, removing the multiplicative amplifier of cond/uncond quantization mismatch.}
    \label{fig:main}
    \vspace{-1.5em}
\end{figure}

The failure mode identified in Sec.~\ref{subsec:stage_behavior} traces to two timestep-agnostic design choices: a uniform distillation objective during training, and a fixed CFG scale during inference. To address them, we present \textbf{DSAQuant}, a \textbf{D}enoising-\textbf{S}tage-\textbf{A}ligned QAT framework with two stage-aware components: \textbf{Denoising-Stage Oriented Supervision} (Sec.~\ref{sec:training state}) and \textbf{Denoising-Stage Gated Guidance} (Sec.~\ref{sec:inference stage}).

\subsection{Training Stage: Denoising-Stage Oriented Supervision}
\label{sec:training state}

The standard target-driven loss used in pretraining of diffusion models is:
\begin{equation}
    \mathcal{L}_{\mathrm{Target}}(t)=\left\|f^q(x_t,c,t) - y_t\right\|_2^2.
    \label{eq:loss_target}
\end{equation}
Instead, traditional QAT paradigms~\citep{he2024efficientdm, huang2026qvgen, li2023qdm} usually train the quantized model by imitating the full-precision model, using distillation loss:

\begin{equation}
    \mathcal{L}_{\mathrm{KD}}(t) =\left\|f^q(x_t,c,t) - f(x_t,c,t)\right\|_2^2.
    \label{eq:loss_kd}
\end{equation}

In early high-noise steps, the noisy latent $x_t$ contains limited information about the clean target $x_1$. In this regime, teacher imitation is useful in two complementary ways. First, constraining the quantized model to follow the full-precision teacher preserves functional similarity in semantic control, spatial layout, and coarse motion, thereby anchoring the global denoising trajectory. Second, the distillation target provides a lower-variance supervision signal than the sample-wise native target: it replaces the uncertain target $y_t$ with the teacher's estimate of the conditional denoising direction, which stabilizes STE-based QAT updates. We analyze this variance-reduction effect in Appendix~\ref{app:gradient_variance}.

However, uniform distillation becomes less suitable in late denoising steps. At this stage, the global structure has largely been formed, and the model mainly refines local appearance, high-frequency textures, object boundaries, and frame-level sharpness. Under low-bit quantization, forcing it to imitate the teacher at every step can therefore waste its limited capacity on matching the teacher residuals rather than minimizing the true diffusion or flow target error. Since late-stage errors directly manifest as blur, texture corruption, and loss of sharpness, the supervision should be oriented toward detail refinement.

We therefore propose \textit{Denoising-Stage Oriented Supervision}, a smooth denoising-stage hybrid objective that preserves teacher distillation in early steps and gradually shifts later steps toward the original diffusion or flow target. Formally, we define the overall training objective as a time-dependent weighted sum of the distillation loss and the native target loss:
\begin{equation}
    \mathcal{L}_{\mathrm{mix}}(t) = \lambda_{\mathrm{KD}}(t) \mathcal{L}_{\mathrm{KD}}(t) + \lambda_{\mathrm{Target}}(t) \mathcal{L}_{\mathrm{Target}}(t)
\end{equation}
We design a normalized exponential scheduler to smoothly govern the transition. The weighting functions are formulated as:
\vspace{-0.5em}
\begin{equation}
    \lambda_{\mathrm{KD}}(t) = \frac{\exp(\alpha (1-t)) - 1}{\exp(\alpha) - 1}, \quad
    \lambda_{\mathrm{Target}}(t) = 1 - \lambda_{\mathrm{KD}}(t)
    \label{eq:weight_target}
\end{equation}
Here, $\alpha > 0$ controls the strength of the exponential decay. By construction, $\lambda_{\mathrm{KD}}(0)=1$ and $\lambda_{\mathrm{Target}}(1)=1$: early steps are fully driven by KD, anchoring layout and stabilizing the STE gradient, while late steps are driven by the target loss, freeing discrete capacity for high-frequency detail.

\subsection{Inference Stage: Denoising-Stage Gated Guidance}
\label{sec:inference stage}
Classifier-free guidance (CFG) is commonly used during inference to improve text-video alignment. We write CFG in the equivalent form
\begin{equation}
    \hat{f}_{\omega}(x_t,c,t)
    =
    f(x_t,\emptyset,t)
    +
    \omega
    \left(
        f(x_t,c,t)-f(x_t,\emptyset,t)
    \right),
    \label{eq:cfg_omega}
\end{equation}
where $\omega$ is the guidance strength.
Let $y_t=y_t(x_1,\epsilon,t)$ denote the sample-wise original diffusion or flow target. The target is shared by the conditional and unconditional branches.
We denote the optimal conditional and unconditional predictors as
\begin{equation}
    m_c(t)=\mathbb{E}[y_t \mid x_t,c],
    \qquad
    m_{\emptyset}(t)=\mathbb{E}[y_t \mid x_t].
    \label{eq:optimal_predictors}
\end{equation}
For the quantized model, we decompose the two predictions as
\begin{equation}
    f_c^q(t)=m_c(t)+\eta_c^q(t),
    \qquad
    f_{\emptyset}^q(t)=m_{\emptyset}(t)+\eta_{\emptyset}^q(t),
    \label{eq:q_decompose}
\end{equation}
where $\eta_c^q(t)$ and $\eta_{\emptyset}^q(t)$ denote quantization-induced deviations from the ideal predictors.

Substituting Eq.~\eqref{eq:q_decompose} into Eq.~\eqref{eq:cfg_omega}, the guided prediction error with respect to the true target becomes
\vspace{-0.5em}
\begin{equation}
    \hat{f}_{\omega}^q(t)-y_t = \underbrace{m_{\emptyset}(t)-y_t}_{\text{target residual}} + \underbrace{\eta_{\emptyset}^q(t)}_{\text{uncond quant error}} + \omega \Big( \underbrace{m_c(t)-m_{\emptyset}(t)}_{\text{ideal guidance direction}} + \underbrace{\eta_c^q(t)-\eta_{\emptyset}^q(t)}_{\text{branch quantization mismatch}} \Big).
\end{equation}

This decomposition shows that CFG amplifies not only the ideal guidance direction, but also the mismatch between the conditional and unconditional quantization errors. In quantized VDMs, the two branches can induce different activation distributions, leading to different clipping and rounding with dynamic activation quantization. Therefore, the quantization errors from the two branches are only weakly correlated; their difference tends to accumulate rather than cancel, and CFG amplifies this mismatch (analysis in Appendix~\ref{app:proof_cfg}).

\begin{wrapfigure}[10]{r}{0.5\textwidth}
  \centering
  \includegraphics[width=0.48\textwidth]{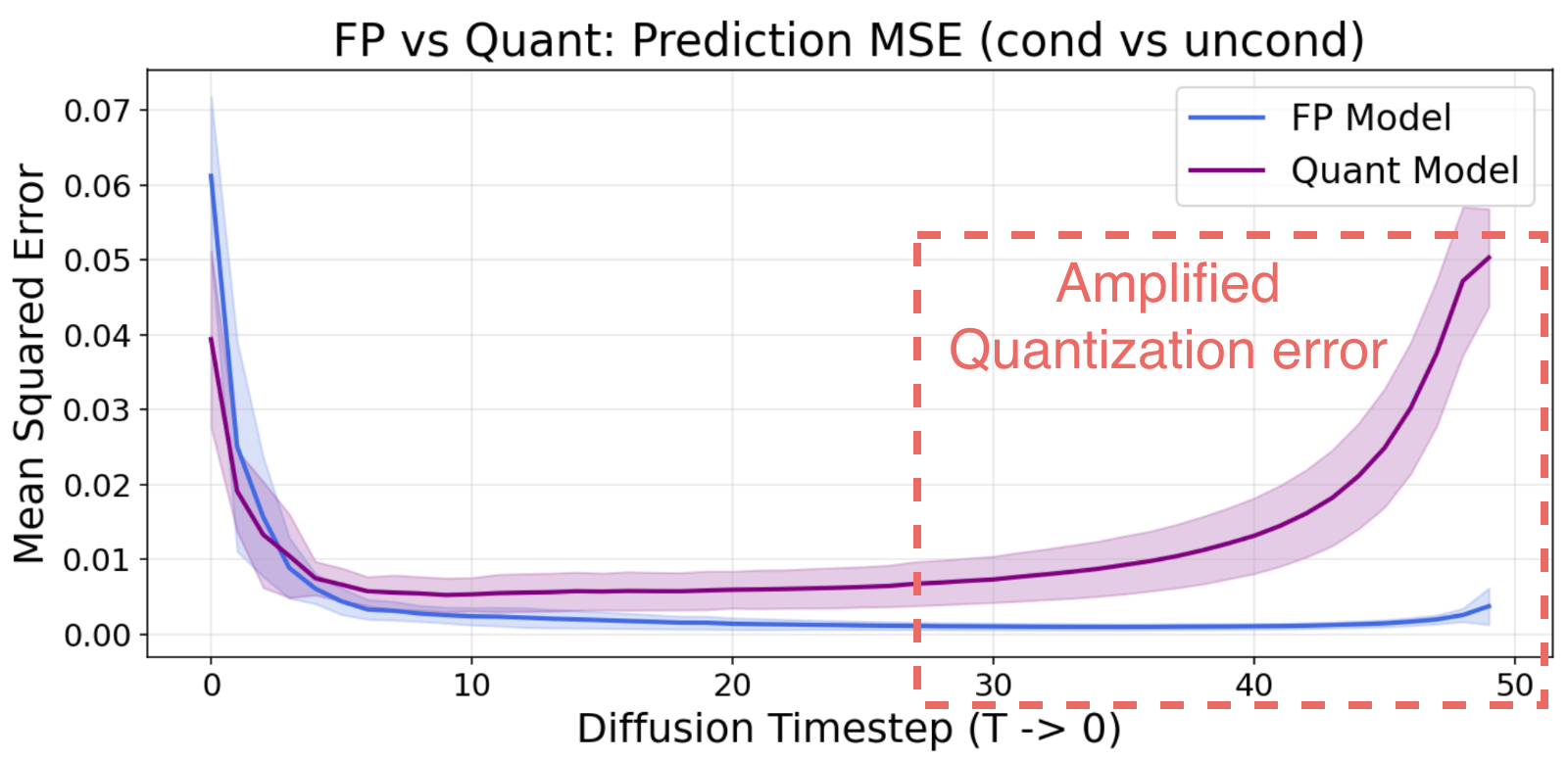}
  \vspace{-0.5em}
  \caption{\textbf{Cond/uncond prediction MSE across denoising timesteps for FP vs.\ quantized models.} While the FP model's MSE stays near zero throughout, the quantized model's MSE rises sharply in late timesteps (boxed region), showing that branch quantization mismatch dominates the (vanishing) ideal guidance signal.}
  \label{fig:cfg-compare}
\end{wrapfigure}

Early denoising benefits from CFG for prompt alignment and global structure formation.
As the denoising process progresses, $x_t$ encapsulates increasingly rich video details, leading to a relative decrease in shared semantic information; consequently, the ideal guidance gradually diminishes. However, as $x_t$ deviates further from the initial Gaussian distribution, the quantization error continuously amplifies. As a result, the branch mismatch paradoxically emerges as the dominant factor in the trailing timesteps, which could lead to high-frequency artifacts. As evidenced in Fig.~\ref{fig:cfg-compare}, the mismatch error rises rapidly in those late timesteps.
We therefore propose a gated guidance scheduler:
\begin{equation}
    \hat{f}_{\omega}^q(t) =
    \begin{cases}
        f_{\emptyset}^q(t) + \omega \big(f_c^q(t) - f_{\emptyset}^q(t)\big), & t \le \tau, \\
        f_c^q(t), & t > \tau.
    \end{cases}
\end{equation}

which removes the amplification of the branch quantization mismatch in the final few steps. (Computationally, the unconditional branch can also be skipped during these late steps, since it is no longer used.)

\begin{table}[t]
  \centering
  \caption{\textbf{Performance comparison across different quantization methods on VBench~\citep{huang2024vbench}.} ``$^*$'' indicates results taken from the QVGen~\citep{huang2026qvgen} paper. ``$^\dagger$'' denotes QVGen's original asymmetric results from the public checkpoint. ``Full Prec.'' denotes the full-precision (BF16) model. Best results are highlighted in \textbf{bold}.}
\vspace{-1.2em}
\renewcommand{\arraystretch}{0.85}
\resizebox{\linewidth}{!}{%
\setlength{\tabcolsep}{2.8pt}
\begin{tabular}[t]{l|c|c|ccccccccc}

\toprule
\multirow{2}{*}{Method} & \multirow{2}{*}{\makecell{\#Bits \\(W/A)}} & \multirow{2}{*}{\makecell{Sym}} & \multirow{2}{*}{\makecell{Imaging\\Quality}$\uparrow$} &  \multirow{2}{*}{\makecell{Aesthetic\\Quality}$\uparrow$} & \multirow{2}{*}{\makecell{Motion\\Smoothness}$\uparrow$} & \multirow{2}{*}{\makecell{Dynamic \\Degree}$\uparrow$} & \multirow{2}{*}{\makecell{Background\\Consistency}$\uparrow$}  & \multirow{2}{*}{\makecell{Subject \\Consistency}$\uparrow$} & \multirow{2}{*}{\makecell{Scene\\Consistency}$\uparrow$} & \multirow{2}{*}{\makecell{Overall\\Consistency}$\uparrow$} & \multirow{2}{*}{\makecell{Average\\$\uparrow$}}\\
 & & & & & & & & & & & \\

\midrule

\multicolumn{12}{c}{\cellcolor[gray]{0.92}\texttt{Wan2.1} $1.3$B ($\texttt{CFG}=5.0, 480p, \texttt{fps}=16$)} \\

\midrule
Full Prec.& 16/16 & -- & 64.22 & 57.85 & 97.25 & 71.11 & 95.72 & 93.34 & 26.07 & 24.70 & 66.28 \\

\midrule
ViDiT-Q$^*$ & 4/6 & $\times$ & 56.24 & 50.18 & 94.81 & 52.43 & 89.67 & 82.53 & 13.45 & 19.58 & 57.36 \\
SVDQuant$^*$ & 4/6 & $\checkmark$ & 58.16 & 51.27 & 97.05 & 49.44 & 93.74 & 91.71 & 14.18 & 23.26 & 59.85 \\
\midrule
LSQ$^*$ & 4/4 & $\times$ & 59.11 & 49.09 & \textbf{98.35} & 71.11 & 92.66 & 91.67 & 10.38 & 18.83 & 61.40 \\
Q-DM$^*$ & 4/4 & $\times$ & 60.40 & 52.50 & 97.22 & 76.67 & 93.37 & 89.26 &  {13.28} &  {21.63} & 63.04 \\
EfficientDM$^*$ & 4/4 & $\times$ &  {60.70} &  {53.57} & 96.18 & 56.39 &  {93.74} &  {91.70} & 11.77 & 21.19 & 60.66 \\

QVGen$^\dagger$ & 4/4 & $\times$ & 59.36 & 54.23 & 96.92 & 73.89 & 93.54 & 90.57 & 11.66 & 23.14 & 62.90 \\

QVGen & 4/4 & $\checkmark$ & 59.88 & 54.92 & 97.18 & 69.72 & 94.58 & 90.95 & 18.05 & 23.17 & 63.56 \\
\rowcolor{blue!5}Ours & 4/4 & $\checkmark$ & \bfseries63.52 & \bfseries58.01 & 96.66 & \bfseries80.00 & \bfseries95.52 & \bfseries92.96 & \bfseries26.54 & \bfseries24.49 & \bfseries67.21 \\

\midrule

\multicolumn{12}{c}{\cellcolor[gray]{0.92}CogVideoX-$2$B ($\texttt{CFG}=6.0, 480p, \texttt{fps}=8$)} \\

\midrule

Full Prec.& 16/16 & -- & 58.21 & 54.37 & 97.42 & 65.27 & 94.62 & 92.54 & 37.07 & 25.10 & 65.57 \\

\midrule
ViDiT-Q$^*$ & 4/6 & $\times$ & 54.72 & 43.01 & 92.18 & 43.22 & 90.76 & 81.02 & 26.25 & 20.41 & 56.45 \\

S2Q-VDiT$^*$ & 4/6 & $\checkmark$ & 53.71 & 52.31 & 98.09 & 36.11 & 96.15 & 93.99 & 34.23 & 24.90 & 61.18 \\
\midrule
LSQ$^*$ & 4/4 & $\times$ & 58.73 & 54.20 & 97.57 & 45.00 & 92.97 & 92.41 & 24.06 & 23.17 & 61.01 \\
Q-DM$^*$ & 4/4 & $\times$ & 54.96 & 52.71 & 98.00 & 48.61 & 93.82 & 91.86 & 28.02 & 23.87 & 61.48 \\
EfficientDM$^*$ & 4/4 & $\times$ & 55.96 & 51.97 & 98.03 & 46.67 & 94.10 & 91.70 & 27.76 & 24.28 & 61.31 \\

QVGen$^\dagger$ & 4/4 & $\times$ & 56.41 & 53.21 & 97.88 & 60.83 & 94.06 & 91.66 & 27.82 & 24.54 & 63.30 \\

QVGen & 4/4 & $\checkmark$ & 56.14 & 52.04 & \bfseries97.50 & \bfseries57.77 & 93.92 & 91.23 & 29.33 & 24.47 & 62.80 \\
\rowcolor{blue!5}Ours & 4/4 & $\checkmark$ & \bfseries59.05 & \bfseries55.01 & 97.49 & 57.22 & \bfseries94.32 & \bfseries92.47 &\bfseries34.26& \bfseries25.54 & \bfseries64.42 \\

\midrule
\multicolumn{12}{c}{\cellcolor[gray]{0.92}\texttt{Wan2.2} $5$B ($\texttt{CFG}=5.0, 720p, \texttt{fps}=16$)} \\

\midrule
Full Prec. & 16/16  & -- & 64.31 & 59.03 & 98.23 & 66.11 & 96.56 & 94.82 & 35.18 & 25.54 & 67.47 \\

\midrule
QVGen & 4/4 & $\checkmark$ & 62.85 & 57.95 & 97.92 & 62.50 & 95.63 & 93.21 & 25.52 & 24.79 & 65.04 \\
\rowcolor{blue!5}Ours & 4/4 & $\checkmark$ & \bfseries 63.46 & \bfseries59.86 & \bfseries98.66 & \bfseries64.17 & \bfseries96.10 & \bfseries95.02 & \bfseries29.08 & \bfseries25.33 & \bfseries66.46 \\

\midrule
\multicolumn{12}{c}{\cellcolor[gray]{0.92}CogVideoX1.5-$5$B ($\texttt{CFG}=6.0, 720p, \texttt{fps}=16$)} \\
\midrule
Full Prec. & 16/16 & -- & 65.36 & 59.42 & 98.39 & 60.83 & 96.57 & 95.00 & 38.43 & 26.29 & 67.54 \\
\midrule
QVGen & 4/4 & $\checkmark$ & 65.70 & 56.58 & 98.81 & 47.50 & 96.58 & 94.05 & 30.42 & 25.23 & 64.35 \\
\rowcolor{blue!5}Ours & 4/4 & $\checkmark$ & \bfseries66.63& \bfseries59.17 & \bfseries98.82 & \bfseries54.17 & \bfseries97.04 & \bfseries94.94 & \bfseries38.17 & \bfseries26.04 & \bfseries 66.87 \\

\bottomrule

\end{tabular}%
}
\label{tab:1.3b}
\vspace{-1.5em}
\end{table}

\section{Experiments}
\begin{table}[t]
  \centering
  \caption{\textbf{Performance comparison on VBench~\citep{huang2024vbench} with augmented prompts under W4A4 and W3A3 settings.} ``$^\dagger$'' denotes QVGen's original asymmetric results from the public checkpoint. ``Full Prec.'' denotes the full-precision (BF16) model. Best results are highlighted in \textbf{bold}.}
\vspace{-1.5em}
\resizebox{\linewidth}{!}{%
\setlength{\tabcolsep}{2.8pt}
\renewcommand{\arraystretch}{0.8}
\begin{tabular}[t]{l|c|c|ccccccccc}
\toprule
\multirow{2}{*}{Method} & \multirow{2}{*}{\makecell{\#Bits \\(W/A)}} & \multirow{2}{*}{\makecell{Sym}} & \multirow{2}{*}{\makecell{Imaging\\Quality}$\uparrow$} &  \multirow{2}{*}{\makecell{Aesthetic\\Quality}$\uparrow$} & \multirow{2}{*}{\makecell{Motion\\Smoothness}$\uparrow$} & \multirow{2}{*}{\makecell{Dynamic \\Degree}$\uparrow$} & \multirow{2}{*}{\makecell{Background\\Consistency}$\uparrow$}  & \multirow{2}{*}{\makecell{Subject \\Consistency}$\uparrow$} & \multirow{2}{*}{\makecell{Scene\\Consistency}$\uparrow$} & \multirow{2}{*}{\makecell{Overall\\Consistency}$\uparrow$} & \multirow{2}{*}{\makecell{Average\\$\uparrow$}}\\
 & & & & & & & & & & & \\
 \midrule
 \multicolumn{12}{c}{\cellcolor[gray]{0.92}\texttt{CogVideoX} $2$B ($\texttt{CFG}=5.0, 480p, \texttt{fps}=16$)} \\
\midrule
Full Prec. & 16/16 & - & 60.18 & 60.71 & 97.21 & 72.22 & 93.66 & 91.40 & 51.10 & 27.22 & 69.21 \\
\midrule
QVGen & 4/4 & $\checkmark$ & 60.53 & 59.87 & 97.15 & \bfseries64.80 & 93.43 & 90.41 & 50.07 & 27.13 & 67.94 \\
\rowcolor{blue!5}Ours & 4/4 & $\checkmark$ & \bfseries61.86 & \bfseries62.37 & \bfseries97.36 & 61.28 & \bfseries94.12 & \bfseries90.73 & \bfseries53.34 & \bfseries27.34 & \bfseries68.55 \\
\midrule
QVGen & 3/3 & $\times$ & 56.61 & 56.75 & 97.17 & \bfseries47.22 & 93.34 & 88.65 & 45.45 & 26.21 & 63.93 \\
\rowcolor{blue!5}Ours & 3/3 & $\times$ & \bfseries59.65 & \bfseries59.98 & \bfseries97.34 & 44.72 & \bfseries93.69 & \bfseries90.39 & \bfseries49.21 & \bfseries26.87 & \bfseries65.23\\
\midrule

\multicolumn{12}{c}{\cellcolor[gray]{0.92}\texttt{Wan2.1} $1.3$B ($\texttt{CFG}=5.0, 480p, \texttt{fps}=16$)} \\
\midrule
Full Prec. & 16/16 & - & 64.73 & 65.12 & 97.99 & 70.28 & 96.69 & 93.62 & 46.05 & 25.77 & 70.00 \\
\midrule
QVGen$^\dagger$ & 4/4 & $\times$ & 62.93 & 63.25 & \bfseries98.30 & 62.77 & 95.75 & 93.13 & 40.45 & 25.04 & 67.71\\
QVGen & 4/4 & $\checkmark$ & 58.42 & 59.50 & 97.21 & 63.06 & 94.56 & 89.43 & 32.50 & 24.14 & 64.85 \\
\rowcolor{blue!5}Ours & 4/4 & $\checkmark$ & \bfseries63.99 & \bfseries64.36 & 97.95 & \bfseries73.33 & \bfseries96.30 & \bfseries93.17 & \bfseries44.10 & \bfseries25.62 & \bfseries69.85 \\
\midrule
QVGen & 3/3 & $\times$ & 58.00 & 57.30 & \bfseries98.32 & 35.55 & 94.63 & 90.24 & 32.00 & 22.57 & 61.08 \\
\rowcolor{blue!5}Ours & 3/3 & $\times$ & \bfseries62.50 & \bfseries62.27 & 97.96 & \bfseries68.05 & \bfseries95.54 & \bfseries92.09 & \bfseries37.73 & \bfseries25.30 & \bfseries67.68 \\

\midrule
\multicolumn{12}{c}{\cellcolor[gray]{0.92}\texttt{Wan2.2} $5$B ($\texttt{CFG}=5.0, 720p, \texttt{fps}=16$)} \\
\midrule
Full Prec. & 16/16 & - & 66.95 & 65.96 & 98.65 & 57.22 & 97.68 & 94.25 & 48.25 & 26.17 & 69.39\\
\midrule
QVGen & 4/4 & $\checkmark$ & 64.84 & 63.68 & 98.49 & 52.22 & 96.75 & 93.78 & 44.68 & 25.43 & 67.48 \\
\rowcolor{blue!5}Ours & 4/4 & $\checkmark$ & \bfseries66.06 & \bfseries66.36 & \bfseries98.86 & \bfseries55.27 & \bfseries97.70 & \bfseries94.75 & \bfseries46.21 & \bfseries25.86 & \bfseries68.88 \\
\midrule
QVGen & 3/3 & $\times$ & 63.44 & 62.36 & 98.40 & 46.11 & 96.57 & 93.04 & 41.48 & 24.34 & 65.72 \\
\rowcolor{blue!5}Ours & 3/3 & $\times$ & \bfseries64.18 & \bfseries65.46 & \bfseries98.74 & \bfseries55.27 & \bfseries97.52 & \bfseries94.15 & \bfseries45.67 & \bfseries25.82 & \bfseries68.35 \\

\midrule
\multicolumn{12}{c}{\cellcolor[gray]{0.92}\texttt{Wan2.1} $14$B ($\texttt{CFG}=5.0, 720p, \texttt{fps}=16$)} \\
\midrule
Full Prec. & 16/16 & - & 67.41 & 65.85 & 97.88 & 73.05 & 97.08 & 92.87 & 45.93 & 26.03 & 70.76 \\
\midrule
QVGen & 4/4 & $\checkmark$ & 66.51 & 65.35 & 97.84 & 62.16 & 96.14 & 92.59 & 41.96 & 25.15 & 68.34 \\
\rowcolor{blue!5}Ours & 4/4 & $\checkmark$ & \bfseries67.60 & \bfseries65.88 & \bfseries98.38 & \bfseries68.89 & \bfseries97.33 & \bfseries93.75 & \bfseries43.69 & \bfseries26.04 & \bfseries70.20 \\
\midrule
QVGen & 3/3 & $\times$ & 64.31 & 64.45 & 97.32 & 58.33 & 95.64 & 92.74 & 38.11 & 25.17 & 67.01 \\
\rowcolor{blue!5}Ours & 3/3 & $\times$ & \bfseries66.35 & \bfseries65.57 & \bfseries98.74 & \bfseries64.44 & \bfseries97.22 & \bfseries94.52 & \bfseries45.96 & \bfseries25.87 & \bfseries69.83 \\

\bottomrule
\end{tabular}%

}
\label{tab:aug vbench1}
\vspace{-0.5em}
\end{table}

\begin{table}[t]
\vspace{-0.5em}
  \small
  \centering
  \caption{\textbf{Performance comparison on VBench-2.0~\citep{zheng2025vbench20}.} ``$^*$'' indicates results taken from the QVGen~\citep{huang2026qvgen} paper. ``$^\dagger$'' denotes QVGen's original asymmetric results from the public checkpoint. ``Full Prec.'' denotes the full-precision (BF16) model. Best results are highlighted in \textbf{bold}.}
\vspace{-1.5em}
\begin{adjustbox}{max width=\linewidth,center}
\renewcommand{\arraystretch}{0.6}
\setlength{\tabcolsep}{2.8pt}
\begin{tabular}[t]{l|c|c|cccccc}
\toprule
Method
& \#Bits (W/A)
& Sym
& Creativity$\uparrow$
& Commonsense$\uparrow$
& Control$\uparrow$
& Human Fidelity$\uparrow$
& Physics$\uparrow$
& Average$\uparrow$\\
 \midrule
\multicolumn{9}{c}{\cellcolor[gray]{0.92}\texttt{Wan2.1} $1.3$B ($\texttt{CFG}=5.0, 480p, \texttt{fps}=16$)} \\
\midrule
Full Prec. & 16/16 & - & 50.0 & 63.9 & 35.2 & 78.7 & 59.7 & 57.5 \\
\midrule
QVGen$^\dagger$ & 4/4 & $\times$ & 43.3 & 61.6 & 31.0 & \bfseries81.1 & 57.2 & 54.8  \\
\rowcolor{blue!5}Ours & 4/4 & $\times$ & \bfseries48.5 & \bfseries63.9 & \bfseries35.7 & 76.2 & \bfseries59.0 & \bfseries56.7\\
QVGen & 4/4 & $\checkmark$ & \bfseries46.5 & 57.5 & 28.0 & 70.1 & 57.2 & 51.9 \\
\rowcolor{blue!5}Ours & 4/4 & $\checkmark$ & 45.0 & \bfseries62.4 & \bfseries35.0 & \bfseries74.0 & \bfseries57.3 & \bfseries54.8 \\
\midrule
QVGen & 3/3 & $\times$ & 37.3 & 51.9 & 23.6 & \bfseries72.3 & 52.6 & 47.5 \\
\rowcolor{blue!5}Ours & 3/3 & $\times$ & \bfseries 43.9 & \bfseries61.5 & \bfseries31.0 & 67.1 & \bfseries58.9 & \bfseries52.5 \\

\midrule
\multicolumn{9}{c}{\cellcolor[gray]{0.92}\texttt{Wan2.1} $14$B ($\texttt{CFG}=5.0, 720p, \texttt{fps}=16$)} \\
\midrule
Full Prec. & 16/16 & - & 55.2 & 64.0 & 37.3 & 81.6 & 62.8 & 60.2 \\
\midrule
QVGen$^*$ & 4/4 & $\times$ & 48.0 & 55.5 & 40.0 & 80.5 & 62.2 & 57.2 \\
QVGen & 4/4 & $\checkmark$ & 52.8 & 59.3 & 37.6 & 79.1 & \bfseries62.4 & 58.2 \\
\rowcolor{blue!5}Ours & 4/4 & $\checkmark$ & \bfseries54.6 & \bfseries63.4 & \bfseries37.8 & \bfseries81.3 & \bfseries62.4 & \bfseries59.9 \\
\midrule
QVGen & 3/3 & $\times$ & 47.2 & 60.7 & 31.2 & 77.3 & 57.6 & 54.8 \\
\rowcolor{blue!5}Ours & 3/3 & $\times$ & \bfseries48.5 & \bfseries62.0 & \bfseries38.3 & \bfseries80.3 & \bfseries58.0 & \bfseries57.4 \\

\bottomrule
\end{tabular}
\end{adjustbox}
\label{tab:vbench2}
\vspace{-1.0em}
\end{table}

\subsection{Implementation details}
\noindent\textbf{Models and Baselines.} Following QVGen~\citep{huang2026qvgen}, we conduct experiments on advanced open-source DMs, including CogVideoX-$2$B, CogVideoX1.5-$5$B~\citep{yang2025cogvideox}, Wan2.1 $1.3$B and Wan2.1 $14$B. We further include Wan2.2 $5$B~\citep{wanteam2025wan}, a model with highly compressed VAE for fast generation. We adopt previous powerful PTQ and QAT methods as baselines: ViDiT-Q~\citep{zhao2025viditq}, SVDQuant~\citep{li2025svdquant}, S2Q-VDIT~\citep{feng2025s}, LSQ~\citep{esser2020learnedstep}, Q-DM~\citep{li2023qdm}, EfficientDM~\citep{he2024efficientdm}, and QVGen~\citep{huang2026qvgen}. We primarily compare our method against QVGen because it's the first and only QAT framework tailored for VDMs, to our knowledge.

\noindent\textbf{Quantization settings.}
Following QVGen~\citep{huang2026qvgen}, we use static per-channel weight quantization and dynamic per-token activation quantization. We use symmetric W4A4 quantization for kernel compatibility~\citep{li2025svdquant, ashkboos2024quarot} and additionally report asymmetric W3A3 results for aggressive low-bit settings. Because QVGen does not release its training data, we reproduce it with the same data, iterations, and quantizers as ours. Following~\citep{li2025efficiencymeetsfidelitynovel, sui2024bitsfusion}, we exclude timestep-conditioned linear layers from quantization because their finite-set inference outputs can be precomputed as LUTs (Appendix~\ref{app:timestep_lut}).

\noindent\textbf{Training.} Following~\citep{huang2025self}, we employ synthesized data based on text prompts from a filtered and LLM-extended version of VidProM~\citep{wang2024vidprom} for training. Detailed settings can be found in Appendix~\ref{appdex:training-setting}.

\noindent\textbf{Evaluation.} 
Our evaluation protocol incorporates multiple benchmarks. Following prior works~\citep{zhao2025viditq, feng2025s, huang2026qvgen}, we evaluate $8$ standard dimensions in VBench~\citep{huang2024vbench}. We additionally utilize VBench-2.0~\citep{zheng2025vbench20} to gauge adherence to \textit{physical laws} and \textit{reasoning}. Furthermore, following \citep{zhu2026causalforcingautoregressivediffusion}, we measure Vision Reward and instruction following using high-motion prompts. Finally, the visual clarity of the videos is quantified via Laplacian Variance (LV).

\vspace{-0.5em}
\subsection{Main results}
\textbf{Comparison with baselines.}
Tab.~\ref{tab:1.3b} presents the evaluation results on the standard VBench, alongside a comparison with existing PTQ and QAT frameworks. Under the 4-bit symmetric quantization setting, our method consistently outperforms QVGen and even surpasses its 4-bit asymmetric counterpart. Notably, on the Wan-1.3B model, our approach achieves a +8.49 improvement in scene consistency and a 3.64 gain in image quality. The consistent performance gains observed on the DDIM-based CogVideo further validate the generalizability of our method.
Additional W3A3 VBench results are provided in Appendix~\ref{app:supp_main_results}.

\textbf{Results on complex prompts and lower bitwidths.}
We further evaluate quantized models under augmented prompts, which contain more complex scenes and instructions. Results are summarized in Tab.~\ref{tab:aug vbench1}. Across all tested \texttt{Wan} backbones, our W4A4 models consistently outperform prior quantization baselines and remain close to the full-precision models.
For example, our method improves the average score over QVGen by $1.40$--$2.14$ points at 4-bit, while reducing the gap to full precision to less than $0.6$ points on all models.
Moreover, our method remains robust under more aggressive 3-bit quantization, achieving $2.63$--$6.60$ higher average scores than QVGen.
Furthermore, we assess our models on VBench-2.0. As shown in Tab.~\ref{tab:vbench2}, on Wan2.1-14B, our method maintains high performance even at 3-bit settings.
These results demonstrate that our method preserves strong generation capability even for challenging prompts and lower bit-width settings. 
Representative qualitative comparisons are shown in Appendix~\ref{appendix:visal_comapre}, with full video comparisons
provided in the supplementary material.

\begin{table}[t]
  \centering
  \caption{\textbf{Component ablation on Wan2.1-1.3B (W4A4, sym).} We evaluate Denoising-Stage Oriented Supervision (mixed loss) and Denoising-Stage Gated Guidance (CFG-drop) independently. ``Base'' denotes the engineering-only quantized baseline with neither component. ``LV'' denotes Laplacian Variance. Best results are highlighted in \textbf{bold}; our full framework is shaded.}
  \label{tab:ablation_wan}
  \vspace{-1.2em}

\begin{adjustbox}{width=\linewidth,center}
\renewcommand{\arraystretch}{0.7}
\begin{tabular}[t]{lcccccccccccc}
\toprule
\multirow{3}{*}{\makecell[c]{Method}} & \multirow{3}{*}{\makecell[c]{Mixed\\loss}} & \multirow{3}{*}{\makecell[c]{CFG\\drop}} & \multirow{3}{*}{\makecell[c]{\#Bits \\(W/A)}} & \multicolumn{6}{c}{\makecell{VBench1$\uparrow$}} & \multicolumn{3}{c}{\makecell{Visual Quality$\uparrow$}} \\
\cmidrule(lr){5-10}\cmidrule(lr){11-13}
 & & & & \makecell{Imaging\\Quality} & \makecell{Aesthetic\\Quality} & \makecell{Dynamic\\Degree} & \makecell{Scene\\Consistency} & \makecell{Overall\\Consistency} & \makecell{Avg} & \makecell{LV} & \makecell{Vision\\Reward} & \makecell{Instruc\\Following} \\
\midrule
\multicolumn{13}{c}{\cellcolor[gray]{0.92}\texttt{Wan} $1.3$B ($\texttt{CFG}=5.0, 480p, \texttt{fps}=16$)} \\
\midrule
FP & -- & -- & 16/16 & 64.22 & 57.85 & 71.11 & 26.07 & 24.70 & 66.3 & 808 & 7.51 & 37.5 \\
FP + CFG-drop & -- & $\checkmark$ & 16/16 & 63.57 & 57.99 & 70.56 & 27.50 & 24.74 & 66.4 & 850\\
\midrule
Base & $\times$ & $\times$ & 4/4 & 62.86 & 56.32 & 71.67 & 22.94 & 24.35 & 65.1 & 674 & 1.59 & 25.8 \\
\midrule
Base + CFG-drop & $\times$ & $\checkmark$ & 4/4 & 61.67 & 56.75 & 77.50 & 25.58 & 24.20 & 66.3 & 842 & 4.51 & 28.9 \\
\midrule
Base + Mixed loss & $\checkmark$ & $\times$ & 4/4 & \bfseries64.06 & 57.47 & 76.67 & 25.16 & \bfseries24.55 & 66.2 & 806 & 4.04 & 29.1 \\
\midrule
\rowcolor{blue!5}Ours & $\checkmark$ & $\checkmark$ & 4/4 & 63.52 & \bfseries58.01 & \bfseries80.00 & \bfseries26.54 & 24.49 & \bfseries67.2 & \bfseries1026 & \bfseries6.47 & \bfseries34.5\\
\bottomrule
\end{tabular}
\end{adjustbox}
\end{table}

\begin{table}[t]
  \centering
  \caption{\textbf{Ablation on the mixing schedule of Denoising-Stage Oriented Supervision} (Wan2.1-1.3B, W4A4, sym). ``linear'' uses a linear schedule; ``exp$\alpha$'' uses the exponential schedule of Eq.~\ref{eq:weight_target} with parameter $\alpha$. CFG-drop is applied by default. Our choice (exp5) is shaded; best results are in \textbf{bold}.}
  \label{tab:ablation_mix}
  \vspace{-1.2em}
\renewcommand{\arraystretch}{0.7}
\begin{adjustbox}{width=\linewidth,center}
\begin{tabular}[t]{ccccccccccccc}
\toprule
\multicolumn{2}{c}{\multirow{2}{*}{\makecell[c]{Method}}} & \multirow{2}{*}{\makecell[c]{\#Bits \\(W/A)}} & \multirow{2}{*}{\makecell[c]{Sym}} & \multicolumn{6}{c}{\makecell{VBench1$\uparrow$}} & \multicolumn{3}{c}{\makecell{Visual Quality$\uparrow$}} \\
\cmidrule(lr){5-10}\cmidrule(lr){11-13}
\multicolumn{2}{c}{} & & & \makecell{Imaging\\Quality} & \makecell{Aesthetic\\Quality} & \makecell{Dynamic\\Degree} & \makecell{Scene\\Consistency} & \makecell{Overall\\Consistency} & \makecell{Avg} & \makecell{LV} & \makecell{Vision\\Reward} & \makecell{Instruc\\Following} \\
\midrule
\multicolumn{13}{c}{\cellcolor[gray]{0.92}\texttt{Wan} $1.3$B ($\texttt{CFG}=5.0, 480p, \texttt{fps}=16$)} \\
\midrule

\multicolumn{2}{c}{FP} & 16/16 & -- & 64.22 & 57.85 & 71.11 & 26.07 & 24.70 & 66.3 & 808 & 7.51 & 37.5 \\

\midrule

\multirow{4}{*}{\makecell[c]{Mix\\loss}} & linear & 4/4 & $\checkmark$ & 62.81 & 57.13 & 70.55 & 20.21 & 23.84 & 65.2 & 862 & 3.80 & 29.7 \\
\cmidrule{2-13}
 & exp1 & 4/4 & $\checkmark$ & 63.42 & 57.03 & 73.61 & 22.60 & 23.85 & 65.9 & 887 & 4.11 & 28.4 \\
\cmidrule{2-13}
 & \cellcolor{blue!5}exp5 & \cellcolor{blue!5}4/4 & \cellcolor{blue!5}$\checkmark$ & \cellcolor{blue!5}\bfseries63.52 & \cellcolor{blue!5}\bfseries58.01 & \cellcolor{blue!5}\bfseries80.00 & \cellcolor{blue!5}\bfseries26.54 & \cellcolor{blue!5}\bfseries24.49 & \cellcolor{blue!5}\bfseries67.2 & \cellcolor{blue!5}\bfseries1026 & \cellcolor{blue!5}\bfseries6.47 & \cellcolor{blue!5}\bfseries34.5 \\
\cmidrule{2-13}
 & exp10 & 4/4 & $\checkmark$ & 63.50 & 57.76 & 74.70 & 18.79 & 23.43 & 65.6 & 785 & 4.19 & 24.8 \\
\bottomrule
\end{tabular}
\end{adjustbox}
\end{table}

\subsection{Ablation study}
To demonstrate the effect of each design, we employ W4A4 quantization on Wan 1.3B for ablation studies. Besides VBench, we also evaluate VisionReward and LV for a comprehensive comparison.

\textbf{Ablation on two key designs.}
First, we evaluate the two proposed components in a 2-by-2 ablation, as summarized in Tab.~\ref{tab:ablation_wan}. Denoising-Stage Oriented Supervision improves perceptual quality by aligning supervision with denoising stages, while Denoising-Stage Gated Guidance (CFG-drop) targets late-step guidance amplification. Combining the two components yields the best overall performance, suggesting that the training-side and inference-side designs are complementary. Representative
qualitative comparisons are provided in the supplementary material.

\textbf{Ablation on schedulers for Denoising-Stage-Oriented Supervision.}
Table~\ref{tab:ablation_mix} compares linear and exponential schedules for balancing distillation loss and target loss.
The exponential schedule with $\alpha=5$ performs best, achieving the highest VBench average and visual quality scores.
This suggests that a moderately sharp decay of the distillation loss better balances teacher guidance at early timesteps and target-loss optimization at later timesteps, while an overly aggressive decay with $\alpha=10$ harms performance.

\begin{wrapfigure}[11]{r}{0.5\textwidth}
  \centering
  \vspace{-4em}
  \includegraphics[width=0.48\textwidth]{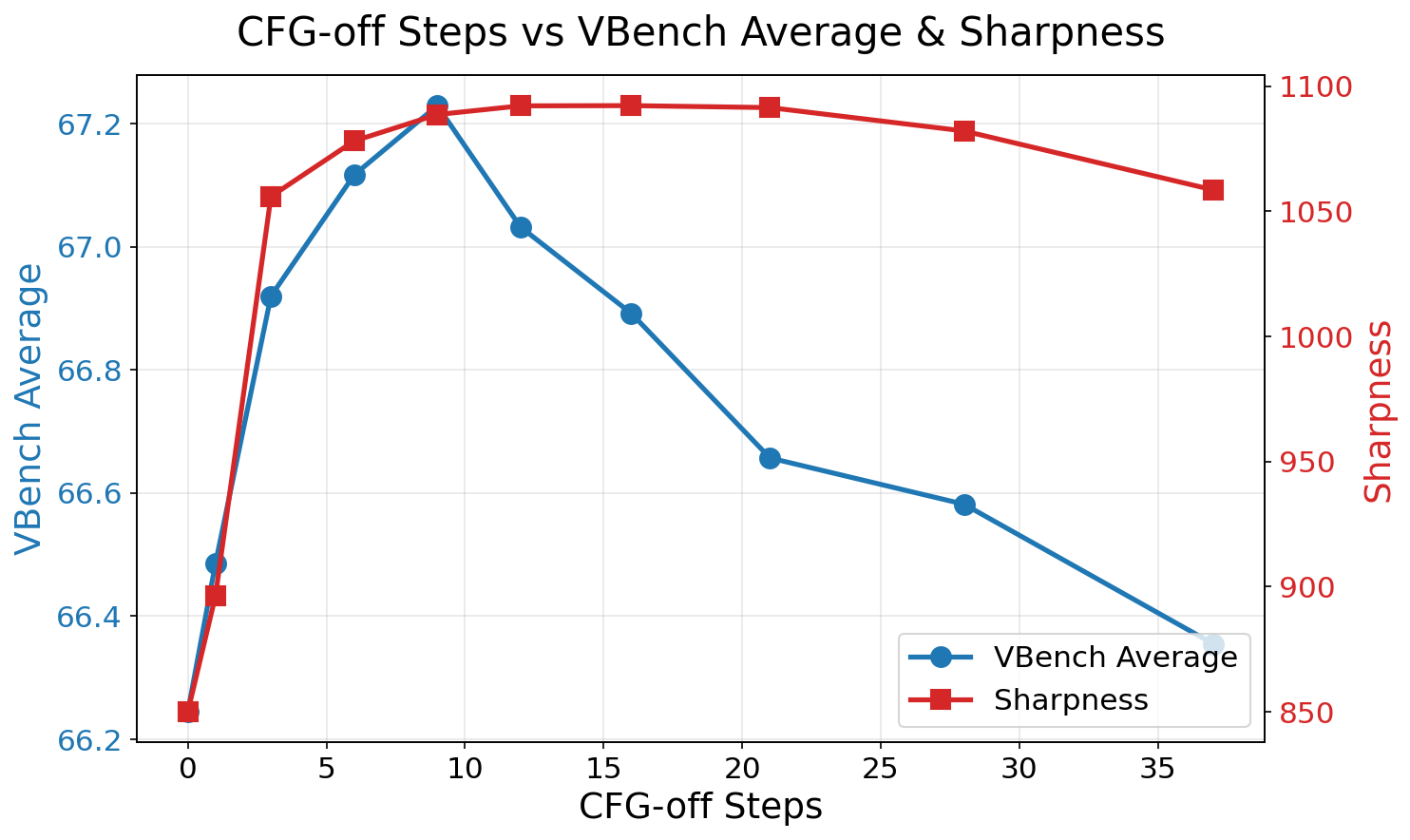}
  \caption{Ablation on the threshold of Denoising-Stage Gated Guidance (CFG-drop).}
  \label{fig:ablation-cfg}
  \vspace{-1.5em}
\end{wrapfigure}

\textbf{Ablation on the threshold of Denoising-Stage-Gated Guidance.}
Fig.~\ref{fig:ablation-cfg} illustrates the impact of the CFG-drop threshold on both visual clarity and VBench scores. We observe that applying CFG-drop even to only the final three timesteps yields a substantial boost in both metrics, effectively suppressing the amplification of quantization errors induced by CFG. However, when CFG-drop is extended beyond ten steps, the VBench score begins to decline, suggesting that premature removal of CFG compromises the semantic integrity of the generated video. Consequently, we set the threshold to nine steps, achieving a near-optimal balance between VBench performance and visual clarity.

\subsection{Efficiency analysis}
\textbf{Training efficiency.}
Table~\ref{training efficiency} reports raw GPU-days together with the GPU type used for each method.
In our training setup, H100 is approximately $2.5\times$--$2.8\times$ faster than H20\footnote{The H100 SXM and H20 are rated at $989$ and $148$ BF16 TFLOPS, respectively, per NVIDIA's official specifications; the practical training-throughput gap is narrower than the $\sim$$6.7\times$ peak ratio because H20's higher HBM bandwidth ($4.0$ vs.\ $3.35$ TB/s) partially compensates the compute deficit on memory-bound operators (attention, optimizer updates, gradient synchronization).}, so after normalizing both methods to H100-equivalent GPU-days, DSAQuant reduces training cost by roughly $2\times$--$7\times$, with the largest gap on \texttt{Wan}~$14$B, indicating favorable scaling to large video diffusion models.

\begin{table}
\small
\renewcommand{\arraystretch}{0.7}
\centering
\caption{\textbf{Training cost on four VDMs, reported in raw GPU-days.} QVGen$^*$ times are taken from the QVGen paper~\citep{huang2026qvgen} (measured on H100); ours are measured on H20 under the settings in Appendix~\ref{appdex:training-setting}. The per-row GPU is shown explicitly to avoid hardware-implicit comparison.}
\begin{tabular}{l|c|cccc}
\toprule
 Method & GPU & CogVideoX-$2$B & \texttt{Wan} $1.3$B & CogVideoX1.5-$5$B & \texttt{Wan} $14$B\\
\midrule
QVGen$^*$ & H100 & 9.44 & 11.11 & $\sim 51$ & $\sim 182$ \\
\rowcolor{blue!5} Ours  & H20 & 6.07 & 8.19 & $\sim 65$ & $\sim 64$\\
\bottomrule
\end{tabular}
\label{training efficiency}
\end{table}

\textbf{Inference efficiency.}
Table~\ref{inference efficiency} shows that our W4A4 model consistently improves inference efficiency on the \texttt{Wan} series.
Compared with FP16, it reduces storage by over $73\%$, peak memory by up to $45.1\%$, and A6000 latency by up to $38.0\%$.
Notably, our quantized \texttt{Wan} $14$B can run on 4090D, while the FP16 model leads to OOM. Moreover, our technique is orthogonal to attention acceleration methods, and combined results are provided in Appendix~\ref{supple: further acceleration}.

\begin{table}
\centering
\small
\renewcommand{\arraystretch}{0.6}
\caption{\textbf{Memory and inference latency on the Wan series, measured under 81 frames at 480p.}}
\vspace{-0.5em}
\begin{tabular}{l|l|llll}
\toprule
Model & Methods
& \makecell[l]{Storage\\(GB)}
& \makecell[l]{Peak inference\\memory (GB)}
& \makecell[l]{A6000\\latency (s)}
& \makecell[l]{4090D\\latency (s)} \\
\midrule
\multirow{2}{*}{Wan 2.1-1.3B}
& FP16 & 2.6 & 8.7 & 313  & 246\\
& Ours w4a4
& 0.7 {\color{blue}(-73.1\%)}
& 7.0 {\color{blue}(-19.5\%)}
& 264 {\color{blue}(-15.7\%)}
& 193 {\color{blue}(-21.5\%)} \\
\midrule
\multirow{2}{*}{Wan 2.1-14B}
& FP16 & 26.6 & 37.7 & 1845 & OOM\\
& Ours w4a4
& 6.7 {\color{blue}(-74.8\%)}
& 20.7 {\color{blue}(-45.1\%)}
& 1268 {\color{blue}(-31.3\%)}
& 930 \\
\midrule
\multirow{2}{*}{Wan 2.2-5B}
& FP16 & 9.5 & 16.9 & 129 & 100 \\
& Ours w4a4
& 2.4 {\color{blue}(-74.7\%)}
& 10.9 {\color{blue}(-35.5\%)}
& 80 {\color{blue}(-38.0\%)}
& 58 {\color{blue}(-42.0\%)} \\
\bottomrule
\end{tabular}
\label{inference efficiency}
\vspace{-0.5em}
\end{table}

\section{Conclusion}
\label{sec:conclusion}

We present DSAQuant, a denoising-stage aligned quantization-aware training framework for video diffusion models. Our key observation is that conventional QAT preserves coarse semantics but disrupts the detail-refinement capability of later denoising steps, resulting in degraded visual fidelity. DSAQuant addresses this issue through Denoising-stage-Oriented Supervision during training and Denoising-stage-Gated Guidance during inference, aligning both supervision and guidance with the stage-wise behavior of video denoising. Experiments demonstrate that DSAQuant improves visual quality, sharpness, and temporal stability while maintaining text-video alignment, highlighting the importance of stage-aware design for high-fidelity VDM quantization.

\begin{ack}
We would like to thank Yichong Lu, Jiahao Wang, Yufeng Yuan, Nan Zhou, Jiahao Shao, and Yudong Jin for their insightful discussions.
\end{ack}


\medskip

\bibliographystyle{plain} 
\bibliography{reference}


\appendix


\newpage
\section{More Related Work}
\label{app:related_work}

\textbf{Quantization for image diffusion models.}
Early diffusion quantization studies mainly focus on image generation. PTQ methods such as Q-Diffusion~\citep{li2023qdiffusion}, PTQD~\citep{he2023ptqd}, and related works~\citep{shang2023posttraining, wang2024accurate} show that diffusion models require timestep-aware calibration because activation distributions vary substantially along the denoising trajectory. Temporal Dynamic Quantization~\citep{so2023temporal}, TFMQ-DM~\citep{huang2024tfmq}, and Temporal Feature Matters~\citep{huang2025temporal} further exploit temporal information to reduce quantization error across denoising steps. For DiT-based image generators, PTQ4DiT~\citep{wu2024ptq4dit} and Q-DiT~\citep{chen2024qdit} study activation outliers and dynamic activation quantization in transformer diffusion backbones, while SVDQuant~\citep{li2025svdquant} isolates difficult outlier components with a low-rank branch. Beyond PTQ, QAT methods such as Q-DM~\citep{li2023qdm} and EfficientDM~\citep{he2024efficientdm} finetune quantized diffusion models with teacher distillation, improving low-bit fidelity compared with calibration-only approaches. More recent image-diffusion quantization frameworks also explore extremely low-bit weights and timestep-related feature handling~\citep{sui2024bitsfusion, li2025efficiencymeetsfidelitynovel}.

\textbf{From image diffusion quantization to video diffusion quantization.}
Although these image-diffusion methods establish strong baselines, directly transferring them to VDMs is non-trivial. Video DiTs process much longer token sequences, making online transformations, rotation-based smoothing, or auxiliary compensation modules more costly at inference time. More importantly, video generation is sensitive not only to single-frame perceptual quality but also to temporal consistency, motion structure, and late-stage detail refinement. Existing VDM quantization methods mainly follow PTQ-style designs. QVD~\citep{tian2024qvd} targets video-specific activation distributions by observing skewed temporal features and inter-channel range disparities, then introduces high temporal discriminability quantization and scattered channel range integration to preserve temporal guidance and improve quantization-level coverage. ViDiT-Q~\citep{zhao2025viditq} extends PTQ to image and video diffusion transformers by analyzing DiT-specific quantization errors and designing a DiT-oriented quantization scheme; its mixed-precision variant further allocates precision to sensitive layers and timesteps. S$^2$Q-VDiT~\citep{feng2025s} focuses on the long-token calibration difficulty of video DiTs, using Hessian-aware salient data selection and attention-guided sparse token distillation to select more informative calibration samples and emphasize influential tokens. QVGen~\citep{huang2026qvgen} introduces QAT to VDMs by adapting the teacher-distillation pipeline used in image diffusion quantization. In contrast, our work shows that this timestep-agnostic QAT objective is mismatched with the stage-wise behavior of video denoising: early steps benefit from teacher anchoring, while late steps require target-oriented detail refinement and reduced CFG-induced quantization-noise amplification.



\section{Gradient Variance Analysis of Early-Stage Distillation}
\label{app:gradient_variance}
This section analyzes the variance-reduction aspect of early-stage distillation. The trajectory-anchoring effect follows from the KD objective itself, which explicitly penalizes deviations between the quantized student and the full-precision teacher.

We use the notation in Sec.~\ref{sec:training state}. For a fixed noisy latent, condition, and timestep, let
\begin{equation}
    z=(x_t,c,t), \qquad s_Q(z)=f^q(x_t,c,t), \qquad J_Q(z)=\nabla_{\theta_Q} f^q(x_t,c,t),
\end{equation}
where $J_Q$ denotes the surrogate Jacobian used by the Straight-Through Estimator (STE). Under squared-error training, the optimal predictor for the native diffusion or flow target is the conditional mean
\begin{equation}
    m(z)=\mathbb{E}\!\left[y_t\mid z\right].
\end{equation}
We decompose the sample-wise target as
\begin{equation}
    y_t=m(z)+\xi_t,
    \qquad
    \mathbb{E}\!\left[\xi_t\mid z\right]=0,
    \qquad
    \operatorname{Cov}\!\left(\xi_t\mid z\right)=\Sigma_t(z).
\end{equation}
Then the conditional expected target loss satisfies
\begin{equation}
    \mathbb{E}\!\left[
        \left\|s_Q(z)-y_t\right\|_2^2
        \mid z
    \right]
    =
    \left\|s_Q(z)-m(z)\right\|_2^2
    +
    \operatorname{Tr}\Sigma_t(z).
\end{equation}
Thus, the sample-wise target loss has the correct population optimum, but its stochastic gradient contains posterior target noise. Specifically, the sample gradient of the target loss in Eq.~\ref{eq:loss_target} is
\begin{equation}
    g_{\mathrm{Target}}(z,y_t)
    =
    \nabla_{\theta_Q}\mathcal{L}_{\mathrm{Target}}(t)
    =
    2 J_Q(z)^\top \left(s_Q(z)-y_t\right).
\end{equation}
and therefore
\begin{equation}
    \operatorname{Cov}\!\left(g_{\mathrm{Target}}\mid z\right)
    =
    4 J_Q(z)^\top
    \Sigma_t(z)
    J_Q(z).
    \label{eq:target_gradient_covariance}
\end{equation}
In early denoising steps, $x_t$ is dominated by noise and contains limited information about the clean video latent $x_1$. For flow, velocity, and data-related targets of the generic form $y_t=a_t x_1+b_t\epsilon$, many plausible clean endpoints remain compatible with the same noisy state. Consequently, the posterior covariance $\Sigma_t(z)$ can be large, and Eq.~\eqref{eq:target_gradient_covariance} shows that this uncertainty is directly injected into the STE gradient. Such high-variance updates are especially harmful in low-bit QAT, where small gradient fluctuations can frequently move latent weights across quantization bin boundaries.

Distillation replaces the sample-wise target with the full-precision teacher prediction. A well-trained teacher can be written as
\begin{equation}
    f(x_t,c,t)=m(z)+\delta(z),
\end{equation}
where $\delta(z)$ denotes the teacher residual with respect to the conditional mean. The sample gradient of the distillation loss in Eq.~\ref{eq:loss_kd} is
\begin{equation}
    g_{\mathrm{KD}}(z)
    =
    \nabla_{\theta_Q}\mathcal{L}_{\mathrm{KD}}(t)
    =
    2 J_Q(z)^\top \left(s_Q(z)-f(x_t,c,t)\right).
\end{equation}
Conditioned on the same $z$, the teacher output is deterministic, so the label-induced conditional covariance is removed:
\begin{equation}
    \operatorname{Cov}\!\left(g_{\mathrm{KD}}\mid z\right)=0.
    \label{eq:kd_gradient_covariance}
\end{equation}
Distillation therefore trades the variance term induced by $\xi_t$ for the teacher residual $\delta(z)$. In early high-noise steps, the posterior target variance is large, while the teacher provides a stable estimate of the conditional denoising direction. KD is therefore preferable as a variance-reduction mechanism: it stabilizes QAT updates and anchors the global denoising trajectory.

This argument also explains why distillation should not dominate all timesteps. As denoising approaches the low-noise refinement regime, the variance-reduction benefit of KD becomes less dominant, while the teacher residual $\delta(z)$ and the limited capacity of the quantized hypothesis space become more important. In this regime, optimizing the native target loss better aligns the quantized model with detail reconstruction instead of forcing it to imitate the full-precision teacher residual. This motivates our denoising-stage schedule, which uses KD in early steps and gradually shifts supervision toward the target loss in later steps.

\section{Two-Stage Analysis of Late-Step CFG Quantization Error}
\label{app:proof_cfg}
We analyze late-step quantization degradation as a two-stage process: a mismatch first forms between the conditional and unconditional branch errors, and CFG then amplifies this mismatch. To isolate quantization effects from denoising-trajectory accumulation, we compare FP and quantized predictions under the same latent state, prompt, and timestep. Define
\begin{equation}
    e_c(t)=f_c^q(t)-f_c(t),
    \qquad
    e_{\emptyset}(t)=f_{\emptyset}^q(t)-f_{\emptyset}(t),
\end{equation}
where $f_c,f_{\emptyset}$ and $f_c^q,f_{\emptyset}^q$ are the FP and quantized conditional and unconditional predictions, respectively.

\paragraph{Stage I: Formation of branch mismatch.}
We first explain why the two branch errors need not cancel under dynamic activation quantization. Consider a dynamic quantizer
\begin{equation}
    Q(h)=s(h)\operatorname{round}\!\left(\frac{h}{s(h)}\right),
    \qquad
    e(h)=Q(h)-h,
\end{equation}
where the scale $s(h)$ is computed from the current activation tensor, token, or channel group. Although the two CFG branches share the same weights and noisy latent, they receive different text conditions. As these signals propagate through cross-attention, normalization, and feed-forward layers, they induce different activation distributions:
\begin{equation}
    h_c^\ell \neq h_{\emptyset}^\ell,
    \qquad
    s_c^\ell=s(h_c^\ell),
    \qquad
    s_{\emptyset}^\ell=s(h_{\emptyset}^\ell).
\end{equation}
In general, $s_c^\ell \neq s_{\emptyset}^\ell$, so the branches are rounded on different quantization grids. Their layer-wise errors are therefore not perfectly aligned, and the accumulated output-level errors $e_c(t)$ and $e_{\emptyset}(t)$ need not cancel. We characterize their alignment by
\begin{equation}
    \rho_t
    =
    \frac{
    \mathbb{E}\!\left[
        \left\langle e_c(t), e_{\emptyset}(t) \right\rangle
    \right]
    }{
    \sqrt{
    \mathbb{E}\|e_c(t)\|_2^2
    \mathbb{E}\|e_{\emptyset}(t)\|_2^2
    }
    }.
\end{equation}
This formulation does not assume independence: the branches may share correlated error because they use the same weights and latent input. The relevant quantity is the residual mismatch after any correlated components cancel.

Let
\begin{equation}
    \sigma_c^2(t)=\mathbb{E}\|e_c(t)\|_2^2,
    \qquad
    \sigma_{\emptyset}^2(t)=\mathbb{E}\|e_{\emptyset}(t)\|_2^2,
\end{equation}
and define the branch mismatch as
\begin{equation}
    \Delta^q(t)=e_c(t)-e_{\emptyset}(t).
\end{equation}
Its expected energy is
\begin{equation}
\begin{aligned}
    \mathbb{E}\|\Delta^q(t)\|_2^2
    &=
    \sigma_c^2(t)
    +
    \sigma_{\emptyset}^2(t)
    -
    2\rho_t\sigma_c(t)\sigma_{\emptyset}(t).
\end{aligned}
\label{eq:branch_mismatch_energy}
\end{equation}
Equation~\eqref{eq:branch_mismatch_energy} separates two causes of a larger late-step mismatch. First, the individual branch errors can increase as increasingly structured late-stage activations become more sensitive to low-bit quantization. Second, weaker cross-branch error correlation reduces cancellation between the two errors. Together, these effects enlarge $\Delta^q(t)$ before CFG is applied.

\paragraph{Stage II: Amplification of branch mismatch by CFG.}
The FP and quantized guided predictions under the same state are
\begin{equation}
    \hat f_\omega(t)=f_{\emptyset}(t)+\omega\big(f_c(t)-f_{\emptyset}(t)\big),
    \qquad
    \hat f_\omega^q(t)=f_{\emptyset}^q(t)+\omega\big(f_c^q(t)-f_{\emptyset}^q(t)\big).
\end{equation}
Their difference gives the final guided quantization error:
\begin{equation}
    \delta_\omega^q(t)
    =\hat f_\omega^q(t)-\hat f_\omega(t)
    =\underbrace{e_{\emptyset}(t)}_{\text{single-branch error}}
    +\omega\underbrace{\big(e_c(t)-e_{\emptyset}(t)\big)}_{\text{branch mismatch }\Delta^q(t)}.
    \label{eq:guided_quant_error}
\end{equation}
Equation~\eqref{eq:guided_quant_error} makes the second stage explicit: after the branch mismatch $\Delta^q(t)$ has formed in Stage I, CFG directly scales its contribution to the final guided error by $\omega$.

This amplification is most harmful in late denoising steps. In this regime, the useful FP guidance direction
\begin{equation}
    f_c(t)-f_{\emptyset}(t)
\end{equation}
becomes relatively weak because the latent already contains rich visual structure, while the branch mismatch formed in Stage I can continue to grow. The amplified term $\omega\Delta^q(t)$ can therefore dominate the useful guidance signal and appear as high-frequency artifacts. Denoising-Stage Gated Guidance removes this multiplier in the final steps by using only the conditional prediction, suppressing Stage II amplification after the useful FP guidance direction has diminished.

\section{General and Quantization-Specific Effects of CFG-Drop}
\label{app:cfg_drop_control}
Late-step CFG suppression can also benefit full-precision diffusion sampling, raising the question of whether CFG-drop is merely a general sampling heuristic. To separate this generic effect from its role in quantized models, we apply an identical CFG-drop schedule to the full-precision model, QVGen, and our engineering-only quantized baseline (Base), while keeping all other inference settings fixed. Table~\ref{tab:cfg_drop_control} reports both the VBench average and Laplacian Variance (LV), where LV measures visual sharpness.

\begin{table}[t]
  \centering
  \caption{\textbf{Identical CFG-drop controls on Wan2.1-1.3B.} We compare each model with and without the same late-step CFG-drop schedule. QVGen and Base use symmetric W4A4 quantization; the full-precision model uses BF16.}
  \label{tab:cfg_drop_control}
  \vspace{-0.8em}
  \renewcommand{\arraystretch}{0.85}
  \begin{adjustbox}{max width=\linewidth,center}
  \begin{tabular}{l|c|ccc|ccc}
  \toprule
  \multirow{2}{*}{Method}
  & \multirow{2}{*}{\makecell{\#Bits \\(W/A)}}
  & \multicolumn{3}{c|}{VBench Avg$\uparrow$}
  & \multicolumn{3}{c}{LV$\uparrow$} \\
  \cmidrule(lr){3-5}\cmidrule(lr){6-8}
  & & w/o drop & w/ drop & $\Delta$ & w/o drop & w/ drop & $\Delta$ \\
  \midrule
  Full Prec. & 16/16 & 66.28 & 66.44 & +0.16 & 808 & 850 & +42 \\
  \midrule
  QVGen & 4/4 & 63.50 & 64.73 & +1.23 & 665 & 856 & +191 \\
  Base & 4/4 & 65.09 & 66.31 & +1.22 & 674 & 842 & +168 \\
  \bottomrule
  \end{tabular}
  \end{adjustbox}
  \vspace{-0.5em}
\end{table}

CFG-drop yields a small improvement for the full-precision model, so we do not attribute its entire effect exclusively to quantization. However, both quantized models receive substantially larger gains in VBench average and, especially, visual sharpness. These controlled results show that the improvement cannot be explained solely by a generic sampling benefit. Instead, CFG-drop is particularly effective under low-bit quantization, consistent with the mechanism in Appendix~\ref{app:proof_cfg}: once the late-stage branch mismatch has formed, removing CFG prevents this quantization-specific error from being further amplified.

\section{Detailed Training Settings}
The detailed training settings including batch size, steps, training frames, and learning rate are provided in Tab.~\ref{tab:training-settings}.
We train our model on H20 GPUs with 140GB memory. By default we use local batch=1 for training, so the global batch size is equal to the number of GPUs we use.

For training data, we use synthetic samples. For Wan2.1-1.3B, we synthesize data from prompts in the filtered and LLM-extended version of VidProM~\citep{wang2024vidprom}, following previous works~\citep{huang2025self}. For Wan2.1-14B, we directly use the synthetic latent data released by TurboDiffusion~\citep{zhang2025turbodiffusion}. For the other models, to reduce data synthesis cost, we decode the TurboDiffusion latents into videos with the Wan2.1 VAE and then re-encode the videos with the corresponding model-specific VAE to obtain latent training data.

\label{appdex:training-setting}
\begin{table}[h]
    \centering
    \caption{Detailed training settings for each model. We set bz=1 for each gpu to evaluate training time.}
    \label{tab:training-settings}
    \begin{adjustbox}{width=\linewidth,center}
    \begin{tabular}{l|c|c|c|c|c|c|c}
    \toprule
    Models
    & \makecell[c]{Batch\\size}
    & \makecell[c]{Total\\steps}
    & \makecell[c]{Warmup\\steps}
    & Lr
    & Frames
    & Resolution
    & \makecell[c]{Time\\(H20 days)} \\
    \midrule
    CogVideoX-2B
    & 32 & 2000 & 200 & 3e-5 & 49 & 480p & 6.07\\

    CogVideoX1.5-5B
    & 32 & 2000 & 200 & 5e-5 & 81 & 720p & $\sim$65 \\

    Wan2.1 1.3B
    & 24 & 2000 & 200 & 3e-5 & 81 & 480p & 8.19\\

    Wan2.1 14B
    & 64 & 1000 & 100 & 3e-5 & 81 & 480p & $\sim$64\\

    Wan2.2 5B
    & 32 & 4000 & 400 & 3e-5 & 81 & 480p & 6.9\\
    \bottomrule
    \end{tabular}
    \end{adjustbox}
\end{table}
\section{Detailed Evaluation Settings}
\label{appdex:evaluation-setting}
Following QVGen~\citep{huang2026qvgen}, we set the random seed according to the process rank at the beginning of evaluation, ensuring that each generated video is sampled from different random noise. We use 128 GPUs for inference on Wan2.1-14B and 32 GPUs for all other models.

\section{Timestep-Conditioned Linear Layers and LUT Precomputation}
\label{app:timestep_lut}
Previous quantization research~\citep{sui2024bitsfusion, li2025efficiencymeetsfidelitynovel, huang2024tfmq} on image diffusion models has demonstrated that pre-computing time features during inference can enhance accuracy without compromising inference efficiency. This is feasible because, during the inference phase, timesteps are discrete integers (e.g., 1,000 distinct values ranging from 0 to 999), forming a finite set. Therefore, we can pre-compute the time features for all timesteps and store them on the CPU. During inference, only the specific feature corresponding to the current timestep is dynamically loaded onto the GPU. Consequently, all timestep-related linear layers can be entirely removed without incurring any additional GPU memory overhead. Below, we detail the procedure for precomputing the Look-Up Table (LUT) for CogVideoX and Wan, alongside a size comparison between the generated LUT and the original linear layers.

We analyze the timestep-conditioned computation in CogVideoX-2B and
Wan2.1-T2V-1.3B. Let the inference scheduler use a fixed set of
denoising timesteps
\begin{equation}
    \mathcal{T}_K = \{t_1, t_2, \ldots, t_K\},
\end{equation}
where we use $K=50$ in the main comparison. Unless otherwise specified,
all storage costs are computed with fp16/bf16 precision, i.e., 2 bytes
per scalar.

\subsection{CogVideoX-2B}

For CogVideoX-2B, the hidden dimension and timestep embedding dimension are
\begin{equation}
    D_c = 30 \times 64 = 1920,
    \qquad
    E_c = 512,
    \qquad
    L_c = 30.
\end{equation}

Given a scheduler timestep $t$, CogVideoX first computes a sinusoidal
timestep embedding and then maps it through a timestep MLP:
\begin{equation}
    p_t = \operatorname{SinEmb}_{D_c}(t) \in \mathbb{R}^{D_c},
\end{equation}
\begin{equation}
    h_t = \operatorname{TEmb}(p_t) \in \mathbb{R}^{E_c}.
\end{equation}

Here $\operatorname{SinEmb}_{D_c}(\cdot)$ is deterministic and has no
trainable linear weight. The trainable timestep MLP can be written as
\begin{equation}
    h_t
    =
    W^{(2)}_t
    \sigma\left(
        W^{(1)}_t p_t + b^{(1)}_t
    \right)
    + b^{(2)}_t,
\end{equation}
where $\sigma(\cdot)$ denotes the activation function.

Each CogVideoX transformer block contains two timestep-conditioned
AdaLN-Zero modules, one before attention and one before the feed-forward
network. For block $\ell \in \{1,\ldots,L_c\}$ and AdaLN index
$r \in \{1,2\}$, the timestep-conditioned modulation vector is
\begin{equation}
    a_{\ell,r}(t)
    =
    W_{\ell,r}
    \sigma(h_t)
    +
    b_{\ell,r}
    \in \mathbb{R}^{6D_c}.
\end{equation}

This vector is split into six per-channel vectors:
\begin{equation}
    a_{\ell,r}(t)
    =
    \left[
    \beta^v_{\ell,r}(t),
    \gamma^v_{\ell,r}(t),
    g^v_{\ell,r}(t),
    \beta^e_{\ell,r}(t),
    \gamma^e_{\ell,r}(t),
    g^e_{\ell,r}(t)
    \right],
\end{equation}
where
\begin{equation}
    \beta^v_{\ell,r}, \gamma^v_{\ell,r}, g^v_{\ell,r}
    \in \mathbb{R}^{D_c},
    \qquad
    \beta^e_{\ell,r}, \gamma^e_{\ell,r}, g^e_{\ell,r}
    \in \mathbb{R}^{D_c}.
\end{equation}

The final output AdaLN produces only shift and scale:
\begin{equation}
    o(t)
    =
    W_o \sigma(h_t) + b_o
    \in \mathbb{R}^{2D_c},
\end{equation}
\begin{equation}
    o(t)
    =
    \left[
    \beta_o(t),
    \gamma_o(t)
    \right],
    \qquad
    \beta_o(t), \gamma_o(t) \in \mathbb{R}^{D_c}.
\end{equation}

Therefore, for a fixed inference timestep set $\mathcal{T}_K$, the
CogVideoX timestep LUT can be defined as
\begin{equation}
    \mathcal{L}_{\mathrm{Cog}}
    =
    \left\{
    a_{\ell,r}(t_s),
    o(t_s)
    \; \middle| \;
    t_s \in \mathcal{T}_K,
    \ell = 1,\ldots,L_c,
    r \in \{1,2\}
    \right\}.
\end{equation}

The number of cached scalar values is
\begin{equation}
    |\mathcal{L}_{\mathrm{Cog}}|
    =
    K
    \left(
        2 L_c \cdot 6D_c + 2D_c
    \right).
\end{equation}

Substituting $K=50$, $L_c=30$, and $D_c=1920$ gives
\begin{equation}
    |\mathcal{L}_{\mathrm{Cog}}|
    =
    50
    \left(
        2 \cdot 30 \cdot 6 \cdot 1920
        +
        2 \cdot 1920
    \right)
    =
    34{,}752{,}000.
\end{equation}

Thus, the fp16/bf16 LUT size is
\begin{equation}
    34{,}752{,}000 \times 2
    =
    69{,}504{,}000 \text{ bytes}
    \approx
    69.50 \text{ MB}
    \approx
    66.28 \text{ MiB}.
\end{equation}

The total number of timestep-related linear parameters in CogVideoX-2B is
\begin{align}
    N_{\mathrm{Cog}}
    &=
    \underbrace{(D_c E_c + E_c) + (E_c^2 + E_c)}_{\text{time embedding MLP}}
    \nonumber \\
    &\quad+
    \underbrace{2 L_c \left((6D_c)E_c + 6D_c\right)}_{\text{block AdaLN linears}}
    \nonumber \\
    &\quad+
    \underbrace{(2D_c)E_c + 2D_c}_{\text{final AdaLN linear}}.
\end{align}

Substituting $D_c=1920$, $E_c=512$, and $L_c=30$:
\begin{equation}
    N_{\mathrm{Cog}}
    =
    357{,}801{,}728.
\end{equation}

Therefore, the fp16/bf16 parameter storage is
\begin{equation}
    357{,}801{,}728 \times 2
    =
    715{,}603{,}456 \text{ bytes}
    \approx
    715.60 \text{ MB}
    \approx
    682.45 \text{ MiB}.
\end{equation}

\subsection{Wan2.1-T2V-1.3B}

For Wan2.1-T2V-1.3B, the hidden dimension, sinusoidal frequency dimension,
and number of transformer layers are
\begin{equation}
    D_w = 1536,
    \qquad
    F_w = 256,
    \qquad
    L_w = 30.
\end{equation}

Given timestep $t$, Wan first computes a sinusoidal embedding:
\begin{equation}
    r_t = \operatorname{SinEmb}_{F_w}(t) \in \mathbb{R}^{F_w}.
\end{equation}

Then the global timestep embedding is computed as
\begin{equation}
    e_t
    =
    W^{(2)}_e
    \sigma
    \left(
        W^{(1)}_e r_t + b^{(1)}_e
    \right)
    +
    b^{(2)}_e
    \in \mathbb{R}^{D_w}.
\end{equation}

Wan then computes a global six-way timestep projection:
\begin{equation}
    e^0_t
    =
    \operatorname{reshape}
    \left(
        W_p \sigma(e_t) + b_p
    \right)
    \in \mathbb{R}^{6 \times D_w}.
\end{equation}

Unlike CogVideoX-2B, Wan does not use a separate large
timestep-conditioned linear layer inside each transformer block. Instead,
each block $\ell$ has a learned modulation tensor
\begin{equation}
    M_\ell \in \mathbb{R}^{6 \times D_w}.
\end{equation}

The block-specific timestep modulation is obtained by a lightweight
element-wise addition:
\begin{equation}
    q_\ell(t)
    =
    M_\ell + e^0_t
    \in \mathbb{R}^{6 \times D_w}.
\end{equation}

This tensor is split into six vectors:
\begin{equation}
    q_\ell(t)
    =
    \left[
    \beta^a_\ell(t),
    \gamma^a_\ell(t),
    g^a_\ell(t),
    \beta^f_\ell(t),
    \gamma^f_\ell(t),
    g^f_\ell(t)
    \right],
\end{equation}
where the superscripts $a$ and $f$ denote the self-attention and
feed-forward branches, respectively.

The final output head also uses the global timestep embedding $e_t$. Let
the head modulation be
\begin{equation}
    M_{\mathrm{head}} \in \mathbb{R}^{2 \times D_w}.
\end{equation}

Then
\begin{equation}
    q_{\mathrm{head}}(t)
    =
    M_{\mathrm{head}} + e_t
    \in \mathbb{R}^{2 \times D_w},
\end{equation}
which is split as
\begin{equation}
    q_{\mathrm{head}}(t)
    =
    \left[
    \beta_{\mathrm{head}}(t),
    \gamma_{\mathrm{head}}(t)
    \right].
\end{equation}


Therefore, for Wan2.1-T2V-1.3B, it is sufficient to cache only the global
timestep quantities:
\begin{equation}
    \mathcal{L}_{\mathrm{Wan}}
    =
    \left\{
    e_{t_s},
    e^0_{t_s}
    \; \middle| \;
    t_s \in \mathcal{T}_K
    \right\}.
\end{equation}

The number of cached scalar values is
\begin{equation}
    |\mathcal{L}_{\mathrm{Wan}}|
    =
    K
    \left(
        D_w + 6D_w
    \right)
    =
    7KD_w.
\end{equation}

Substituting $K=50$ and $D_w=1536$ gives
\begin{equation}
    |\mathcal{L}_{\mathrm{Wan}}|
    =
    50 \times 7 \times 1536
    =
    537{,}600.
\end{equation}

Thus, the fp16/bf16 LUT size is
\begin{equation}
    537{,}600 \times 2
    =
    1{,}075{,}200 \text{ bytes}
    \approx
    1.075 \text{ MB}
    \approx
    1.025 \text{ MiB}.
\end{equation}

The total number of timestep-related linear parameters in Wan2.1-T2V-1.3B is
\begin{align}
    N_{\mathrm{Wan}}
    &=
    \underbrace{(F_w D_w + D_w) + (D_w^2 + D_w)}_{\text{time embedding MLP}}
    \nonumber \\
    &\quad+
    \underbrace{(6D_w)D_w + 6D_w}_{\text{global time projection}}.
\end{align}

Substituting $F_w=256$ and $D_w=1536$:
\begin{equation}
    N_{\mathrm{Wan}}
    =
    16{,}920{,}576.
\end{equation}

Therefore, the fp16/bf16 parameter storage is
\begin{equation}
    16{,}920{,}576 \times 2
    =
    33{,}841{,}152 \text{ bytes}
    \approx
    33.84 \text{ MB}
    \approx
    32.27 \text{ MiB}.
\end{equation}

\subsection{Why LUT Precomputation is Feasible}

For both models, the cached quantities are deterministic functions of the
discrete scheduler timestep and fixed model weights:
\begin{equation}
    \mathcal{L}(t)
    =
    f_\theta(t),
    \qquad
    t \in \mathcal{T}_K,
\end{equation}
where $\theta$ denotes the frozen parameters of the timestep-conditioned
linear layers.

Importantly, these quantities do not depend on the latent tokens, text
tokens, attention activations, or feed-forward activations:
\begin{equation}
    \frac{\partial \mathcal{L}(t)}{\partial X} = 0,
    \qquad
    \frac{\partial \mathcal{L}(t)}{\partial C} = 0.
\end{equation}

Thus, given a fixed inference scheduler, the timestep-conditioned linear
outputs can be precomputed once before denoising:
\begin{equation}
    \left\{
    f_\theta(t_1),
    f_\theta(t_2),
    \ldots,
    f_\theta(t_K)
    \right\},
\end{equation}
and reused throughout the denoising process by indexing the LUT with the
current timestep.

This removes online execution of the timestep-related linear layers while
preserving the original LayerNorm, affine modulation, attention,
feed-forward, and residual-gating computation paths.

\subsection{Storage Comparison}
The total storage cost comparison is summarized in Tab.~\ref{tab:timestep_lut_storage}.

\begin{table}[t]
\centering
\caption{Comparison of timestep-related linear parameters and 50-step LUT size.}
\label{tab:timestep_lut_storage}
\resizebox{\linewidth}{!}{
\begin{tabular}{lccclc}
\toprule
Model &
Time-related linear scope &
Params &
Weight size &
50-step LUT content &
LUT size \\
\midrule
CogVideoX-2B &
\begin{tabular}{c}
time embedding MLP \\
+ block AdaLN linears \\
+ final AdaLN linear
\end{tabular}
&
357,801,728
&
715.60 MB / 682.45 MiB
&
\begin{tabular}{c}
all block AdaLN parameters \\
+ final AdaLN parameters
\end{tabular}
&
69.50 MB / 66.28 MiB
\\
\midrule
Wan2.1-T2V-1.3B &
\begin{tabular}{c}
time embedding MLP \\
+ global time projection
\end{tabular}
&
16,920,576
&
33.84 MB / 32.27 MiB
&
\begin{tabular}{c}
global $e_t \in \mathbb{R}^{1536}$ \\
and $e^0_t \in \mathbb{R}^{6 \times 1536}$
\end{tabular}
&
1.075 MB / 1.025 MiB
\\
\bottomrule
\end{tabular}
}
\end{table}

\section{Supplementary Results for Main Experiments}
\label{app:supp_main_results}
This section provides supplementary VBench results, summarized in Tab.~\ref{tab:supp_main_vbench}, for the main comparison in Tab.~\ref{tab:1.3b}. We focus on the more aggressive W3A3 setting for Wan2.1-1.3B and Wan2.2-5B, complementing the main-table results with per-dimension scores under standard prompts.

\begin{table}[t]
  \centering
  \caption{Supplementary table for VBench results in Tab.~\ref{tab:1.3b}.}
  \label{tab:supp_main_vbench}
\begin{adjustbox}{width=\linewidth,center}
\begin{tabular}[t]{lcccccccccc}
\toprule
\multirow{3}{*}{\makecell[c]{Method}} & \multirow{3}{*}{\makecell[c]{\#Bits \\(W/A)}} & \multicolumn{9}{c}{\makecell{VBench1$\uparrow$}} \\
\cmidrule(lr){3-11}
 & & \makecell{Imaging\\Quality} & \makecell{Aesthetic\\Quality} & \makecell{Motion\\Smoothness} & \makecell{Dynamic\\Degree} & \makecell{Background\\Consistency} & \makecell{Subject\\Consistency} & \makecell{Scene\\Consistency} & \makecell{Overall\\Consistency} & \makecell{Avg} \\
\midrule
\multicolumn{11}{c}{\cellcolor[gray]{0.92}\texttt{Wan2.1} $1.3$B ($\texttt{CFG}=5.0, 480p, \texttt{fps}=16$)} \\
\midrule
FP & 16/16 & 64.22 & 57.85 & 97.25 & 71.11 & 95.72 & 93.35 & 26.07 & 24.70 & 66.28 \\
\midrule
QVGen & 3/3 & 59.48 & 50.33 & 97.91 & 42.50 & 94.78 & 89.96 & 11.68 & 20.32 & 58.37 \\
Ours & 3/3 & 61.90 & 54.56 & 96.41 & 77.77 & 94.46 & 90.88 & 22.46 & 24.11 & 65.32\\
\midrule
\multicolumn{11}{c}{\cellcolor[gray]{0.92}\texttt{Wan2.2} $5$B ($\texttt{CFG}=5.0, 720p, \texttt{fps}=16$)} \\
\midrule
QVGen & 3/3 & 60.65 & 54.95 & 97.97 & 55.0 & 95.08 & 92.43 & 19.56 & 23.49 & 62.39 \\
Ours & 3/3 & 61.22 & 58.53 & 98.50 & 60.00 & 95.74 & 94.92 & 24.26 & 25.27 & 64.81 \\

\bottomrule
\end{tabular}
\end{adjustbox}
\end{table}

\section{Full VBench Results For Ablation Studies}
\label{supple_ablation}
Due to page constraints in the main text, the ablation study section only reports five sensitive metrics from VBench. The comprehensive evaluation results across all metrics are summarized in Tabs.~\ref{tab:supp_ablation_wan} and~\ref{tab:supp_ablation_mix}.

\begin{table}[t]
  \centering
  \caption{Supplementary table for ablation study in Tab.~\ref{tab:ablation_wan}.}
  \label{tab:supp_ablation_wan}
\begin{adjustbox}{width=\linewidth,center}
\begin{tabular}[t]{lcccccccccccc}
\toprule
\multirow{2}{*}{\makecell[c]{Method}} & \multirow{2}{*}{\makecell[c]{Mixed\\loss}} & \multirow{2}{*}{\makecell[c]{CFG\\drop}} & \multirow{2}{*}{\makecell[c]{\#Bits \\(W/A)}} & \multicolumn{9}{c}{\makecell{VBench1$\uparrow$}} \\
\cmidrule(lr){5-13}
 & & & & \makecell{Imaging\\Quality} & \makecell{Aesthetic\\Quality} & \makecell{Motion\\Smoothness} & \makecell{Dynamic\\Degree} & \makecell{Background\\Consistency} & \makecell{Subject\\Consistency} & \makecell{Scene\\Consistency} & \makecell{Overall\\Consistency} & \makecell{Avg} \\
\midrule
\multicolumn{13}{c}{\cellcolor[gray]{0.92}\texttt{Wan} $1.3$B ($\texttt{CFG}=5.0, 480p, \texttt{fps}=16$)} \\
\midrule
FP & -- & -- & 16/16 & 64.22 & 57.85 & 97.25 & 71.11 & 95.72 & 93.35 & 26.07 & 24.70 & 66.28 \\
\midrule
Base & $\times$ & $\times$ & 4/4 & 62.86 & 56.32 & 96.23 & 71.67 & 94.68 & 91.69 & 22.94 & 24.35 & 65.09 \\
\midrule
Base + CFG-drop & $\times$ & $\checkmark$ & 4/4 & 61.67 & 56.75 & 96.57 & 77.50 & 96.57 & 92.57 & 25.58 & 24.20 & 66.31 \\
\midrule
Base + Mixed loss & $\checkmark$ & $\times$ & 4/4 & 64.06 & 57.47 & 96.07 & 76.67 & 94.45 & 91.52 & 25.16 & 24.55 & 66.24 \\
\midrule
\rowcolor{blue!5}Ours & $\checkmark$ & $\checkmark$ & 4/4 & 63.52 & 58.01 & 96.67 & 80.00 & 95.52 & 92.96 & 26.54 & 24.49 & 67.20 \\
\bottomrule
\end{tabular}
\end{adjustbox}
\end{table}

\begin{table}[t]
  \centering
  \caption{Supplementary table for ablation study in Tab.~\ref{tab:ablation_mix}.}
  \label{tab:supp_ablation_mix}
\begin{adjustbox}{width=\linewidth,center}
\begin{tabular}[t]{lcccccccccc}
\toprule
\multirow{2}{*}{\makecell[c]{Method}} & \multirow{2}{*}{\makecell[c]{\#Bits \\(W/A)}} & \multicolumn{9}{c}{\makecell{VBench1$\uparrow$}} \\
\cmidrule(lr){3-11}
 & & \makecell{Imaging\\Quality} & \makecell{Aesthetic\\Quality} & \makecell{Motion\\Smoothness} & \makecell{Dynamic\\Degree} & \makecell{Background\\Consistency} & \makecell{Subject\\Consistency} & \makecell{Scene\\Consistency} & \makecell{Overall\\Consistency} & \makecell{Avg} \\
\midrule
\multicolumn{11}{c}{\cellcolor[gray]{0.92}\texttt{Wan} $1.3$B ($\texttt{CFG}=5.0, 480p, \texttt{fps}=16$)} \\
\midrule
FP & 16/16 & 64.22 & 57.85 & 97.25 & 71.11 & 95.72 & 93.35 & 26.07 & 24.70 & 66.28 \\
\midrule
linear & 4/4 &62.8 & 57.1 & 97.80 & 70.55 & 95.88 & 93.73 & 20.22 & 23.84 & 65.24\\
\midrule
exp1 & 4/4 & 63.42 & 57.03 & 97.52 & 73.61 & 95.71 & 93.79 & 22.60 & 23.85 & 65.94\\
\midrule
\rowcolor{blue!5} exp5& 4/4 & 63.53 & 58.01 & 96.67 & 80.00 & 95.52 & 92.96 & 26.54 & 24.49 & 67.20 \\
\midrule
exp10 & 4/4 & 63.50 & 57.71 & 97.70 & 74.72 & 95.77 & 93.49 & 18.79 & 23.43 & 65.64 \\
\bottomrule
\end{tabular}
\end{adjustbox}
\end{table}

\section{Further inference acceleration}
\label{supple: further acceleration}
We further integrate SageAttention to accelerate inference. The results are summarized in Tab.~\ref{tab:sage+w4a4}.
\begin{table}[h]
    \centering
    \caption{Inference latency when combining W4A4 quantization with SageAttention. All results are measured under 81 frames at 480p.}
    \begin{tabular}{l|l|c}
    \toprule
    Model & Methods & 4090D latency(s) \\
    \midrule
    \multirow{2}{*}{Wan 2.1-1.3B} & Ours & 193\\
    & Ours + SageAttention & 140 \\
    \midrule
    \multirow{2}{*}{Wan 2.1-14B} & Ours & 930 \\
    & Ours + SageAttention & 698 \\
    \midrule
    \multirow{2}{*}{Wan 2.2-5B} & Ours & 58\\
    & Ours + SageAttention & 53 \\
    \bottomrule
    \end{tabular}
    \label{tab:sage+w4a4}
\end{table}
\section{Visual Comparison}
\label{appendix:visal_comapre}
\begin{figure}
    \centering
    \includegraphics[width=0.9\linewidth]{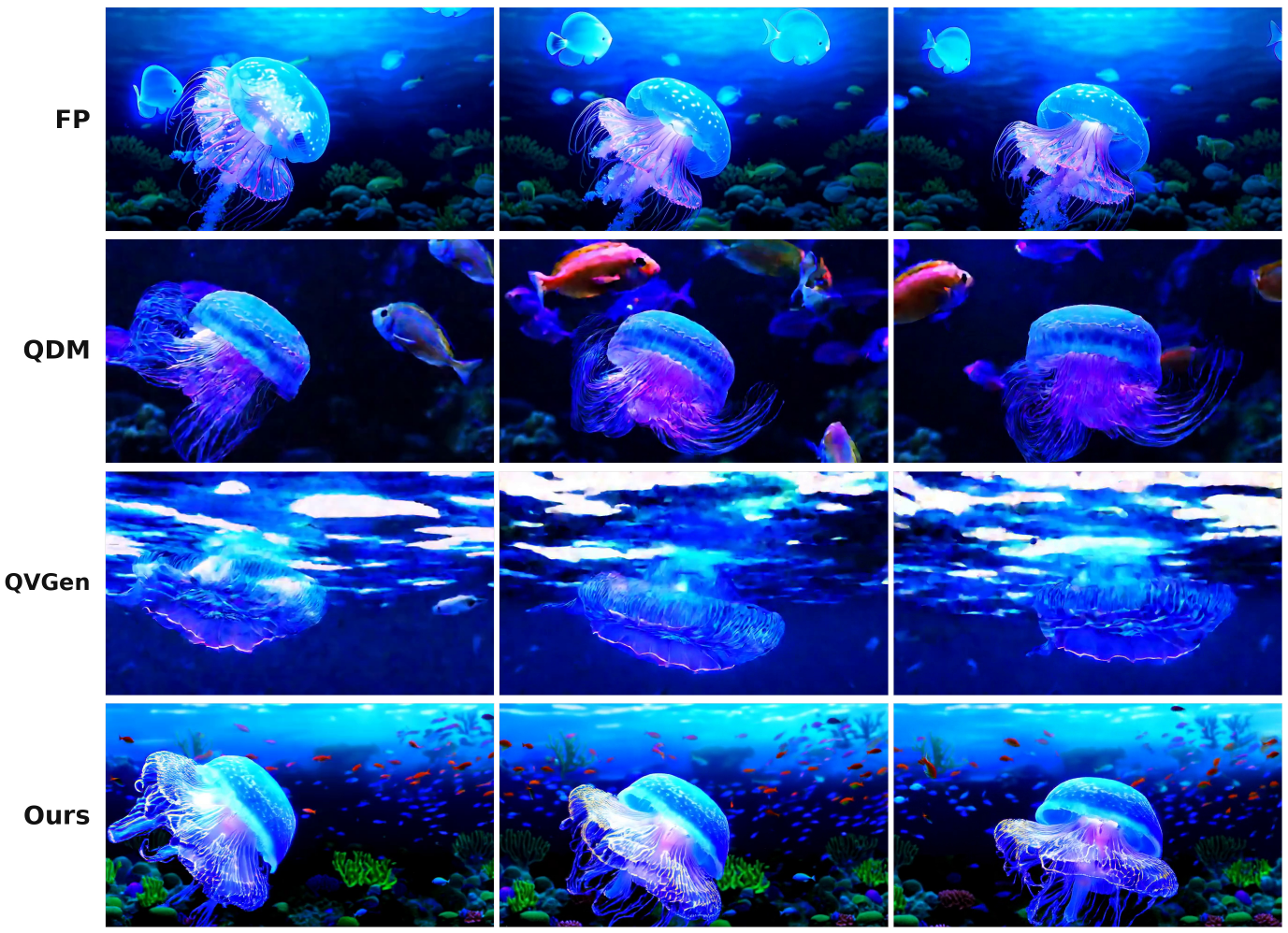}
    \caption{\small 4bit quantization on Wan2.1 1.3b model with prompt:"CG animation digital art, a majestic jellyfish floating gracefully through the oceanic depths. The jellyfish has intricate patterns on its translucent body, with vibrant hues of blue, green, and purple. Its bioluminescent tentacles emit a soft, mesmerizing glow, creating an ethereal underwater landscape. The tentacles sway gently as the jellyfish glides, illuminating the surrounding water with a captivating light show. The ocean background is filled with schools of colorful fish and drifting coral reefs. The jellyfish is surrounded by a serene, tranquil atmosphere. Soft, ambient ocean sounds play in the background. Low-angle, slow-motion shot focusing on the jellyfish and its glowing tentacles."}
    \label{fig_wan1b_4bit}
\end{figure}

\begin{figure}
    \centering
    \includegraphics[width=0.9\linewidth]{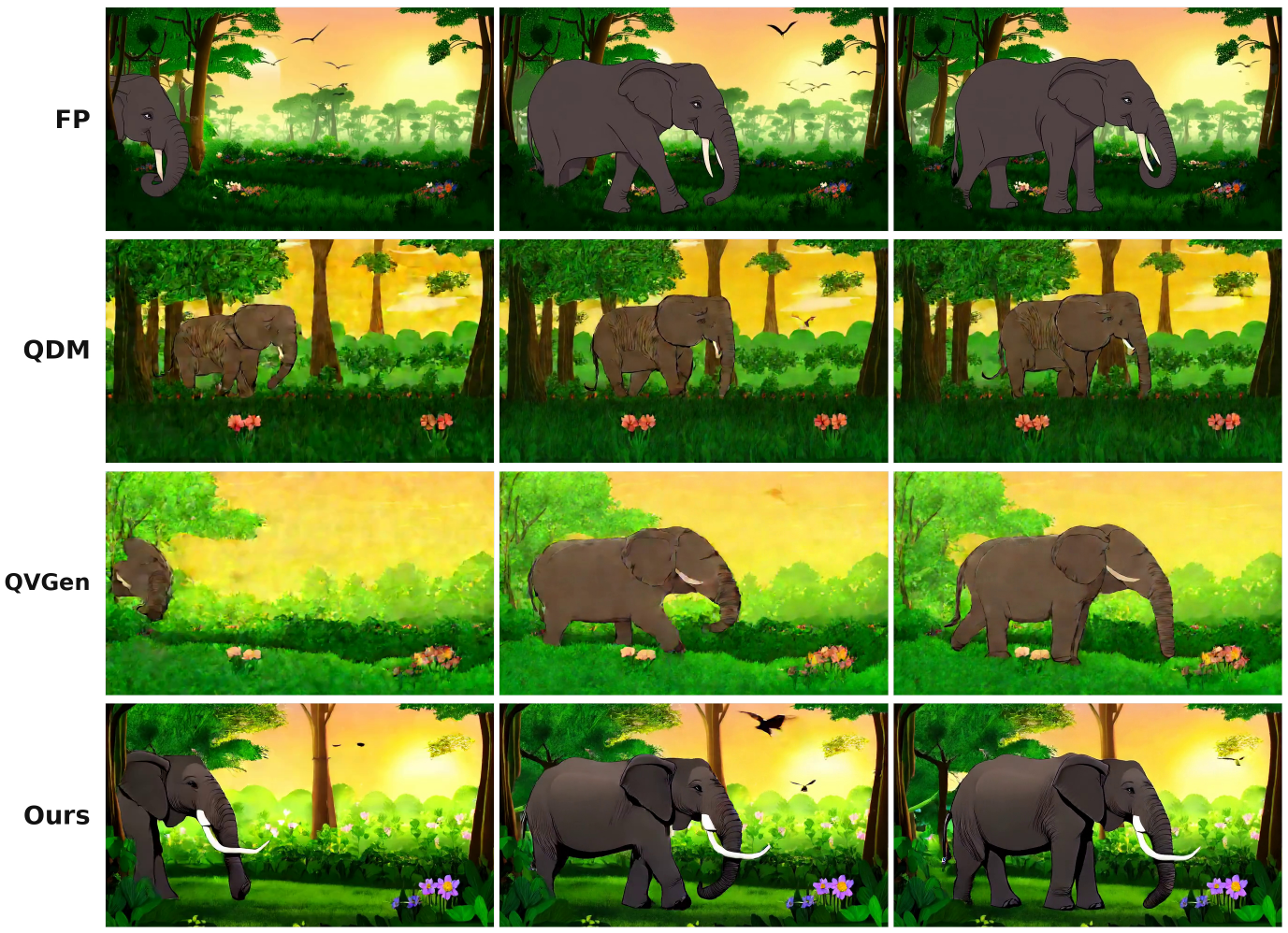}
    \caption{\small 3bit quantization on Wan2.1 1.3b model with prompt:"CG animation digital art, an African elephant taking a peaceful walk through a lush green forest at dawn. The elephant has a grayish-brown coat with soft folds and wrinkles, and gentle eyes. It walks slowly, with its trunk raised slightly, sniffing the air. The forest is filled with tall trees and vibrant greenery, with colorful flowers blooming. Birds fly overhead, chirping melodiously. The sky is a beautiful orange hue as the sun rises. The elephant moves gracefully, with each step careful and deliberate. The background features intricate foliage and subtle shadows. Soft lighting creates a warm and serene atmosphere. Low-angle shot from behind, focusing on the elephant's tranquil expression and movement."}
    \label{fig_wan1b_3bit}
\end{figure}

\begin{figure}
    \centering
    \includegraphics[width=0.9\linewidth]{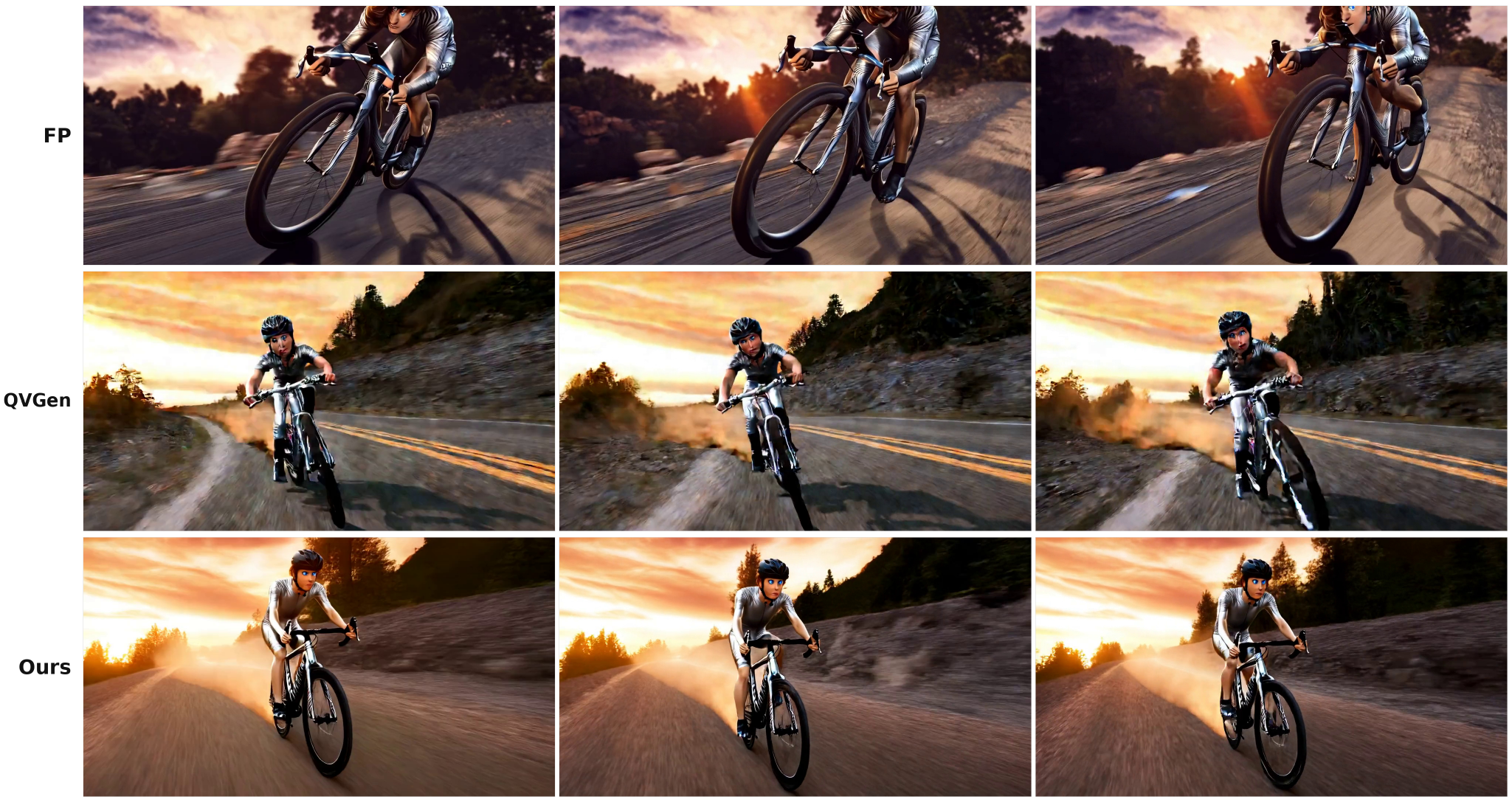}
    \caption{\small  4bit quantization on Wan2.2 5b model with prompt:"CG animation digital art, a sleek racing bicycle speeding down a winding mountain road. The bike is a deep metallic silver color with intricate patterns etched along its frame. It accelerates rapidly, the rider gripping the handlebars tightly with focused determination. The rider has short, wavy brown hair and piercing blue eyes, wearing a tight-fitting black helmet and a snugly fitting silver jersey. They lean forward slightly, their arms pumping hard to maintain momentum. The road is rugged, covered in loose gravel and rocks, with trees lining the sides. The sun sets behind them, casting dramatic shadows. The background features a sunset sky with wispy clouds and hints of orange and pink hues. The bike's wheels spin furiously as it gains speed, creating a whirlwind of dust and debris. The scene captures the adrenaline rush of a thrilling bicycle race. Close-up, low-angle view."}
    \label{fig_wan5b_4bit}
\end{figure}

\begin{figure}
    \centering
    \includegraphics[width=0.9\linewidth]{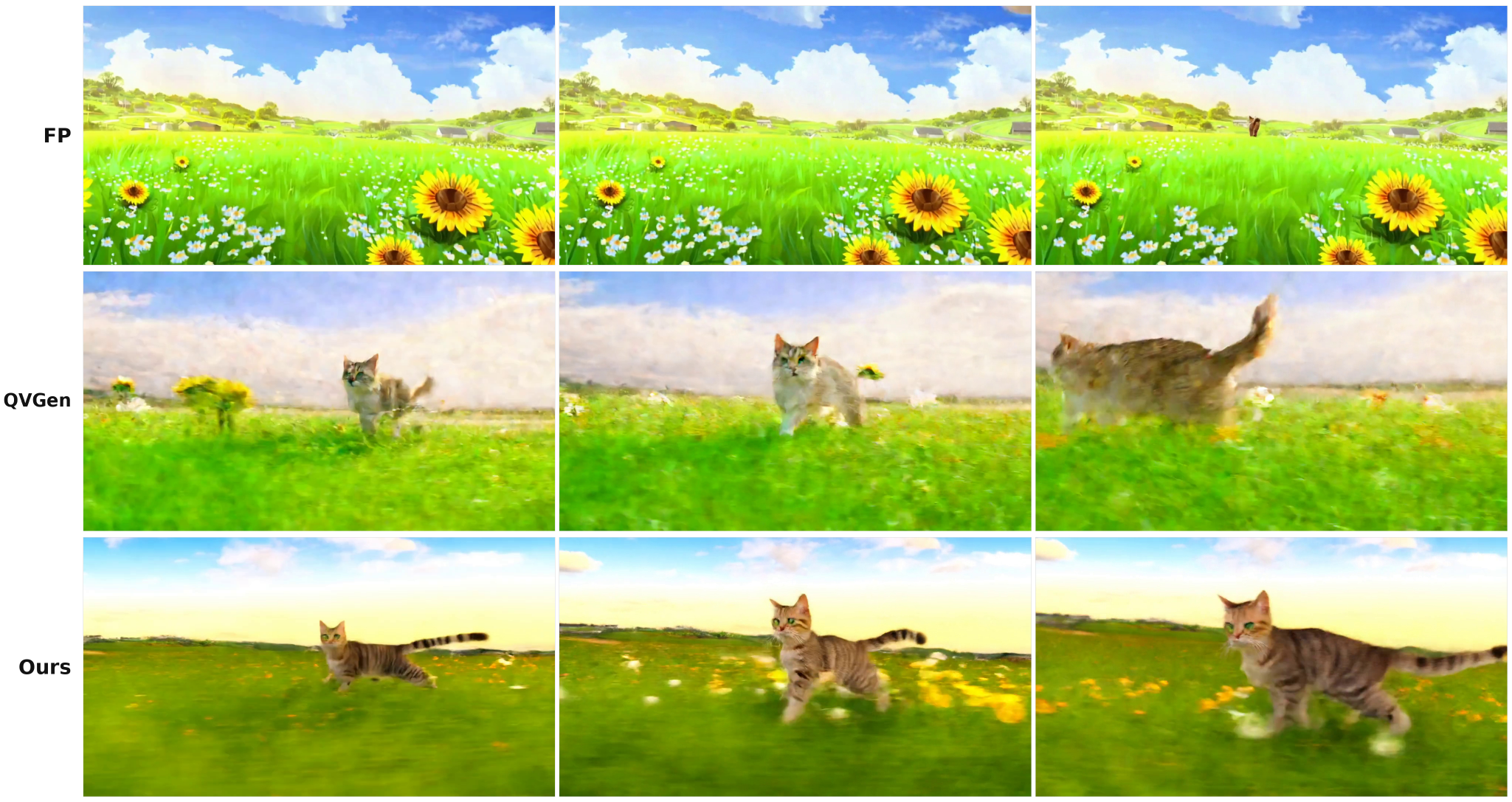}
    \caption{\small  3bit quantization on Wan2.2 5b model with prompt:"A playful feline sprinting joyfully across a lush green meadow dotted with wildflowers. The cat has sleek fur, expressive green eyes, and a fluffy tail that wags excitedly as it bounds forward. The meadow stretches out behind it, with vibrant sunflowers and buttercups swaying gently in the breeze. The sky above is a bright azure, filled with fluffy white clouds. The cat's joyful run is captured from a dynamic low-angle perspective, showcasing its agility and boundless energy. The scene is bathed in warm golden light, enhancing the cat's lively demeanor. Grass and petals trail behind the cat as it dashes towards the horizon. The background features a serene rural landscape, with small cottages and winding country roads visible in the distance. The overall composition is energetic and full of life, perfectly capturing the essence of a cat running happily."}
    \label{fig_wan5b_3bit}
\end{figure}

\begin{figure}
    \centering
    \includegraphics[width=0.9\linewidth]{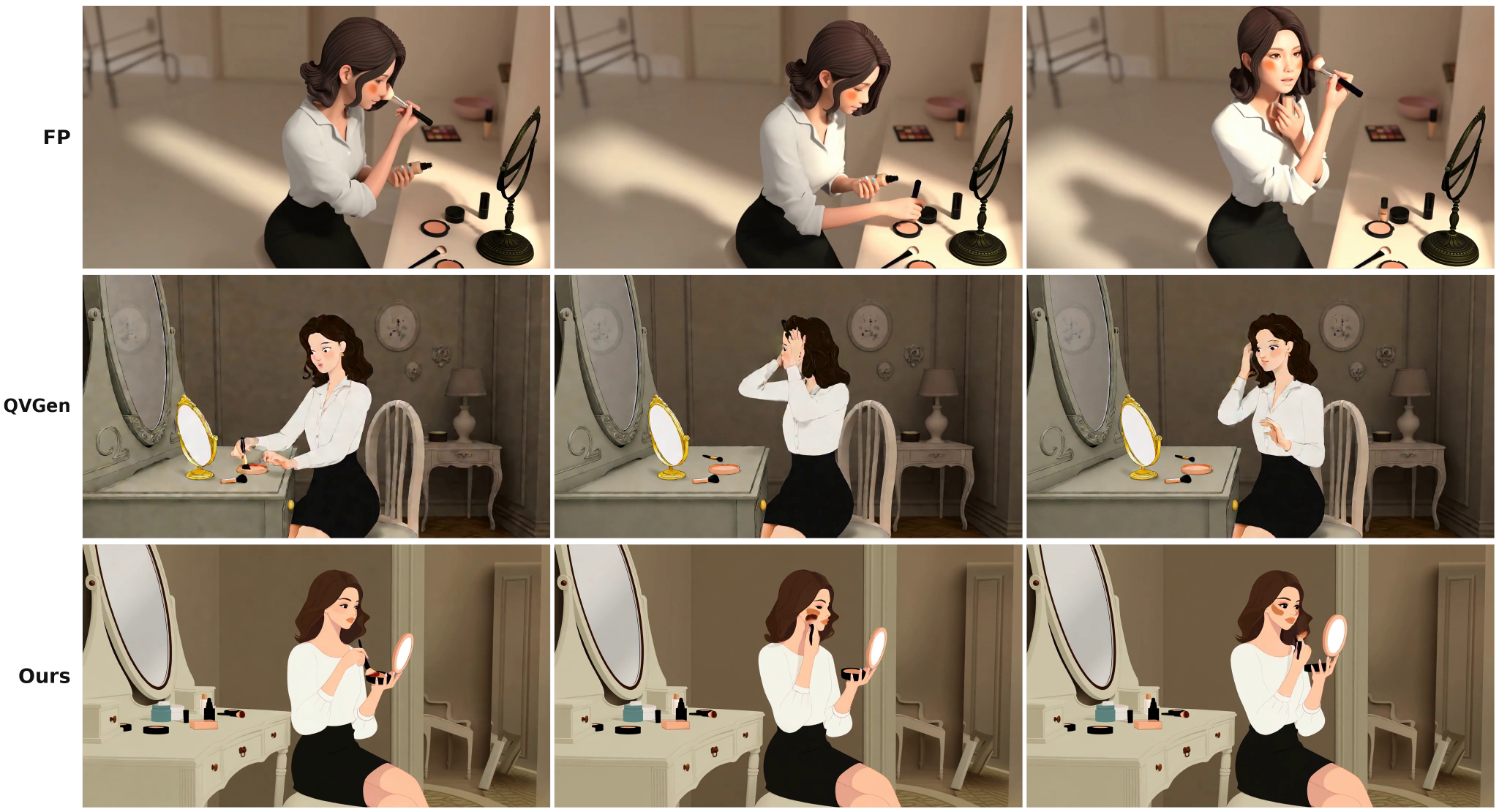}
    \caption{\small  4bit quantization on Wan2.1 14b model with prompt:"CG animation concept art, a young woman with natural beauty applying makeup in the morning. She is wearing a simple white blouse and black pencil skirt. Her hair is styled in loose waves, framing her face. She is sitting at a vanity table with soft lighting illuminating her work. She is using a compact mirror to apply foundation, concealer, and blush. Her fingers move gracefully as she blends and applies products. She pauses occasionally to check her reflection and adjust her technique. The background features a minimalist room with a few scattered items and a vintage vanity set. Soft, gentle brush strokes and subtle motions. Low-angle shot from above, focusing on her hands and facial expressions."}
    \label{fig_wan14b_4bit}
\end{figure}

We provide representative qualitative comparisons across three Wan family
backbones: Wan2.1-1.3B, Wan2.1-14B, and Wan2.2-5B. All samples are generated with the same prompts and inference settings as the quantitative evaluation, allowing a direct comparison between DSAQuant and prior QAT baselines. These examples complement the numerical results by highlighting differences in text-video alignment, texture fidelity, local detail preservation, and quantization-induced artifacts. Specifically, Figs.~\ref{fig_wan1b_4bit} and~\ref{fig_wan1b_3bit} show W4A4 and W3A3 results on Wan2.1-1.3B,
Figs.~\ref{fig_wan5b_4bit} and~\ref{fig_wan5b_3bit} show results on Wan2.2-5B, and Figs.~\ref{fig_wan14b_4bit} and~\ref{fig_wan14b_3bit} show results on Wan2.1-14B. Full video comparisons are provided in the supplementary material.

\begin{figure}
    \centering
    \includegraphics[width=0.9\linewidth]{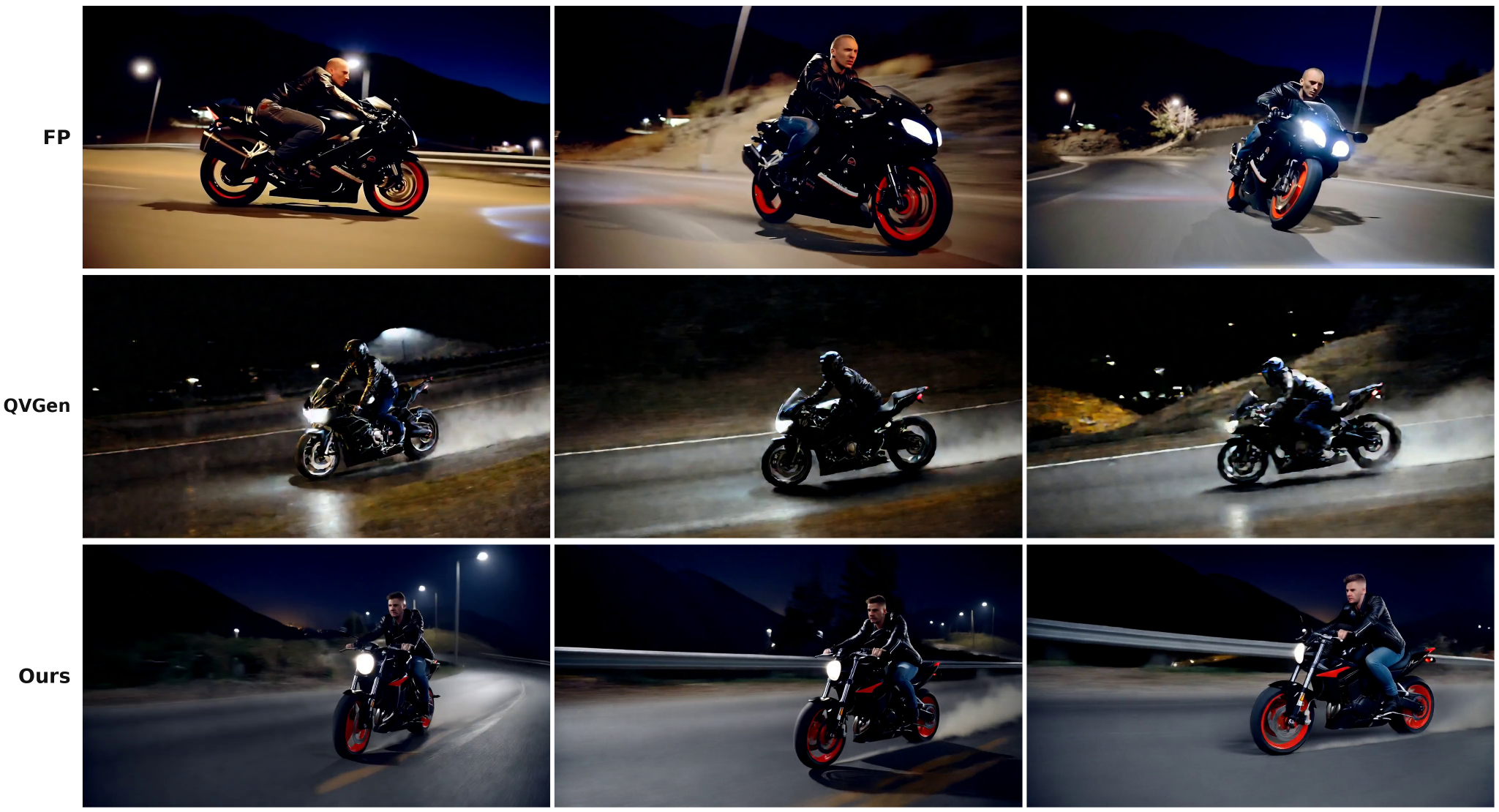}
    \caption{\small  3bit quantization on Wan2.1 14b model with prompt:"A thrilling motorcycle speeding down a winding mountain road at night. The sleek black motorcycle with bright red accents accelerates rapidly, leaving behind a trail of dust. The rider, a muscular man with short cropped hair, wears a black leather jacket and jeans. His helmet is off, revealing a determined and focused expression. He grips the handlebars tightly, leaning slightly into the turn. The motorcycle's engine roars as it gains momentum, reflecting the intense speed and power. The background is a dimly lit mountain landscape, with flickering streetlights and twinkling stars. The scene captures the adrenaline rush and raw energy of the motorcycle's acceleration. Nighttime low-angle shot from the side."}
    \label{fig_wan14b_3bit}
\end{figure}

\section{Limitations, Broader impacts}
\label{appendix:limitation}
\textbf{Limitations.} Our experiments are currently conducted on bidirectional video diffusion models. Recent advances in autoregressive video generation show strong potential for real-time generation, and evaluating DSAQuant on such models is an important direction for future work.

\textbf{Broader impacts.} This work improves the efficiency of video diffusion models through quantization-aware training. By reducing memory and computational costs while preserving visual fidelity, DSAQuant can make high-quality video generation more accessible and practical on resource-constrained hardware. This may benefit creative production, education, assistive applications, and interactive media.
However, efficient video generation also raises potential risks. Lowering the deployment cost of high-fidelity VDMs may make it easier to generate misleading synthetic videos, impersonation content, privacy-violating media, or other harmful outputs. Efficiency gains may also increase overall usage, partially offsetting the environmental benefits of reduced per-sample computation. We therefore encourage the deployment of such models together with safeguards such as watermarking, provenance tracking, content filtering, safety classifiers, and clear disclosure of synthetic content.
\section{LLM Usage}
\label{Appendix:llm_usage}
In this paper, Large Language Models (LLMs) were used to assist with polishing the text and formatting the tables.

\end{document}